\documentclass{article}

\usepackage{PRIMEarxiv}
\usepackage{amsmath,amsfonts}
\usepackage{algorithmic,algorithm}
\usepackage{array}
\usepackage[caption=false,font=normalsize,labelfont=sf,textfont=sf]{subfig}
\usepackage{textcomp}
\usepackage{stfloats}
\usepackage{url}
\usepackage{verbatim}
\usepackage{graphicx, graphics}
\def\BibTeX{{\rm B\kern-.05em{\sc i\kern-.025em b}\kern-.08em
    T\kern-.1667em\lower.7ex\hbox{E}\kern-.125emX}}
\usepackage{balance}
\usepackage{multirow}
\usepackage{booktabs}
\usepackage{orcidlink}
\usepackage{subcaption}

\begin{document}
\title{RadiusFPS: Efficient Farthest Point Sampling on CPUs and GPUs via Spherical Voxel Pruning}



\author{
    Ziyang Yu\\
    School of Computing \\
    Institute of Science Tokyo\\
    Tokyo \\
    \texttt{yu.ziyang.2001@gmail.com}
    \And
    Xiang Li \\
    School of Computing \\
    Institute of Science Tokyo \\
    Tokyo \\
    \texttt{652022230010@smail.nju.edu.cn}
    \And
    Qiong Chang \\
    School of Computing \\
    Institute of Science Tokyo \\
    Tokyo \\
    \texttt{q.chang@c.titech.ac.jp}
    \And
    Jun Miyazaki \\
    School of Computing \\
    Institute of Science Tokyo \\
    Tokyo\\
    \texttt{miyazaki@comp.isct.ac.jp}
}

\maketitle

\begin{abstract}

Point clouds are a primary sensory representation for robotic perception, underpinning LiDAR-based autonomous driving, simultaneous localization and mapping (SLAM), and navigation. Within these pipelines, Farthest Point Sampling (FPS) is the most well-known downsampling operator, as its uniform coverage preserves the geometric structure on which downstream perception relies. However, the large time complexity of classical FPS scales poorly with the million-point-per-second rates of modern 3D sensors, making it a dominant latency bottleneck that conflicts with the real-time and limited onboard compute budgets of robotic systems.
Therefore, we propose RadiusFPS, an FPS acceleration framework based on spherical voxel pruning that preserves the standard FPS update rule under the same initialization and tie-breaking policy. By indexing the point cloud with spherical voxels, RadiusFPS derives a conservative geometric bound that prunes redundant distance computations in each iteration, complemented by a coordinate-wise point-skip test that removes residual updates.
We further introduce RadiusFPS-G, a warp-level GPU implementation that fuses voxel selection, pruning, and distance update into memory-coalesced kernels, eliminating costly global-memory round-trips. On indoor (S3DIS, ScanNet) and outdoor LiDAR (SemanticKITTI) benchmarks, RadiusFPS-G attains up to ~2.5× speedup over GPU-based FPS and matches or exceeds QuickFPS among the evaluated methods while using roughly half its GPU memory, with comparable segmentation accuracy. When coupled with the learning-based FastPoint sampler, the resulting pipeline achieves the fastest End-to-End inference among all evaluated configurations. These properties make high-quality FPS-style sampling practical for latency- and memory-constrained robotic vision.


\end{abstract}


\section{Introduction}

Point clouds have become a fundamental sensory representation in robotic vision, where robots must perceive, interpret, and interact with complex three-dimensional environments. Modern 3D sensors, including LiDAR, RGB-D cameras, and depth cameras, provide dense geometric measurements together with attributes such as color, reflectance, and intensity, enabling robots to estimate scene structure, localize themselves, detect obstacles, and reason about objects and free space. As a result, point cloud processing has become essential to a wide range of robotic applications, including autonomous driving, simultaneous localization and mapping (SLAM), navigation, inspection, and human-robot interaction.
Driven by rapid advances in sensing hardware, the spatial resolution and acquisition rate of point cloud sensors have increased substantially in recent years. For example, contemporary mainstream LiDAR sensors, typically featuring 128 lines, can generate over 1,000,000 points per second, representing a five-fold increase compared with sensors from just five years ago \cite{lidarsensor}. While such dense measurements provide richer geometric details and improve the robustness of robotic perception, they also impose severe computational pressure on downstream algorithms, leading to excessive memory consumption and prohibitive latency. This challenge is particularly critical for robotic systems, where perception modules must operate under strict real-time constraints and limited onboard computational resources. Therefore, sampling a representative subset of points is a basic yet significant task in 3D robotic vision. Effective point cloud sampling can reduce computational overhead while preserving essential geometric structures, making it a key technique for efficient and reliable perception in large-scale robotic systems.

Among point cloud sampling strategies, Farthest Point Sampling (FPS) has become a common sampling strategy for point cloud tasks. Unlike heuristic alternatives such as random or grid sampling, FPS produces a near-uniform spatial distribution that, for the same number of sampled points, retains finer geometric structure and broader scene coverage. This property has made FPS a core component of the perception backbones widely deployed on robotic platforms, spanning point cloud segmentation~\cite{pointmetabase,pointvector,pointnet++,pointransformer, ptv1,ptv2,pointcept,randla}, object detection~\cite{pointrcnn,IASSD,3dssd,3dstd}, classification~\cite{pointmamba,pointbert,pointmae,3dctn,pointgn}, etc.
However, despite its central role, FPS is inherently sequential and compute-intensive, and in practice becomes the dominant latency bottleneck—limiting its applicability precisely in the large-scale, real-time settings that robotic perception demands.

\begin{figure}[t]
\centering

\includegraphics[width=0.48\linewidth]{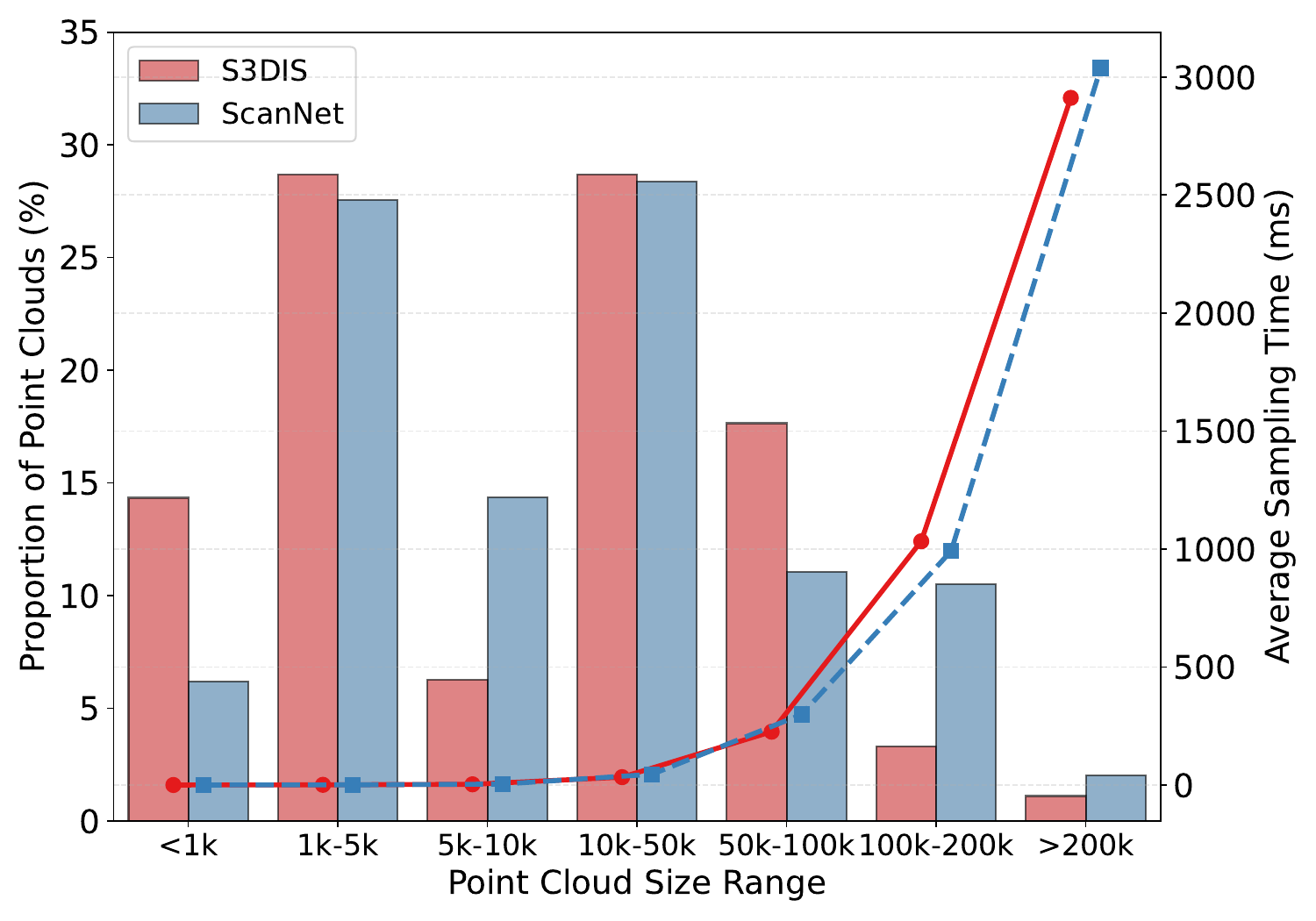}
\includegraphics[width=0.48\linewidth]{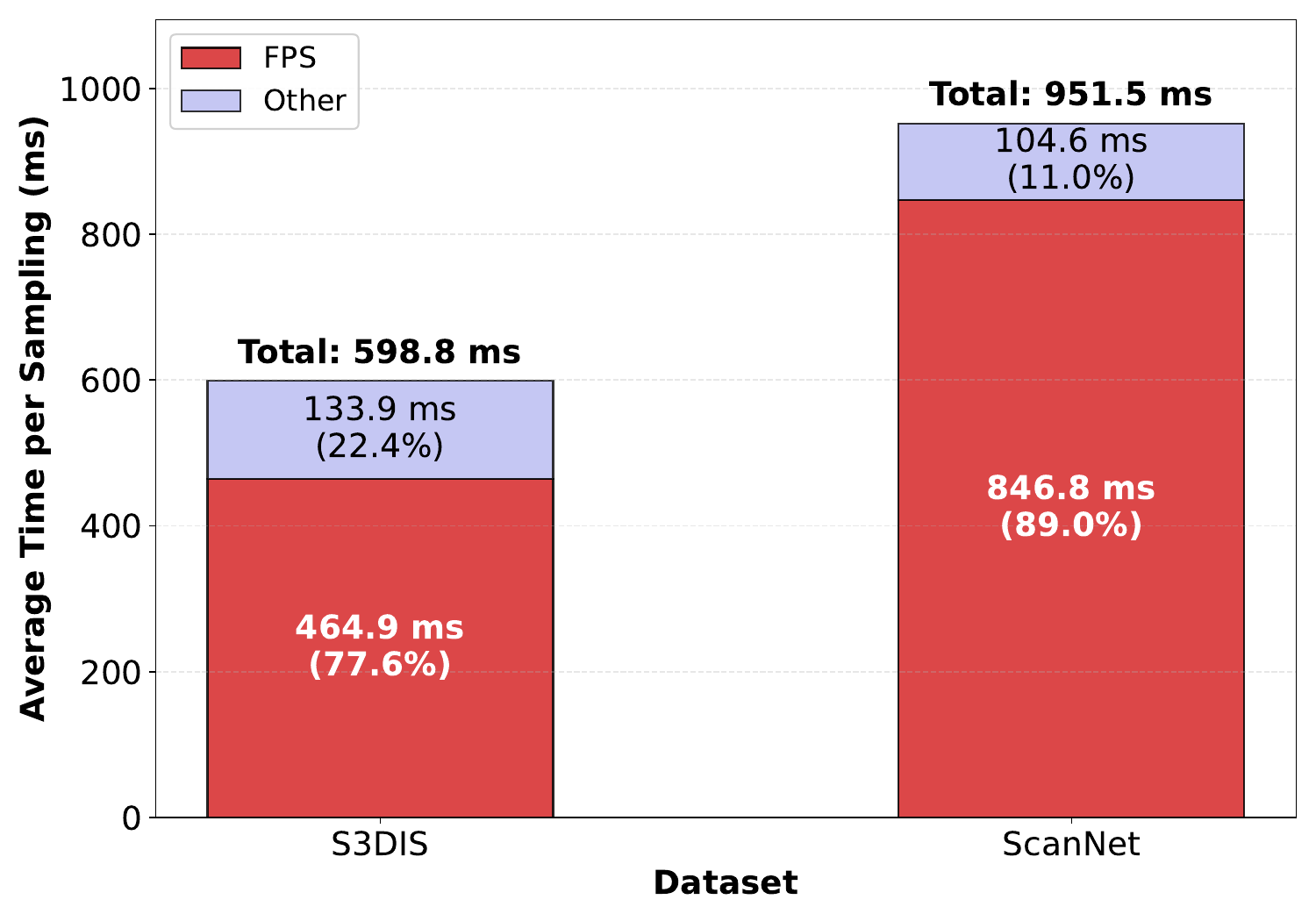}

\caption{Efficiency analysis of FPS on the S3DIS and ScanNet datasets.
Top: point cloud size distribution and sampling latency. The histograms show the proportion of point clouds in different size ranges for S3DIS and ScanNet, while the overlaid curves indicate the average sampling time for each range. The sampling latency increases sharply as the point cloud size grows, especially beyond $50\mathrm{k}$ points.
Bottom: average sampling latency breakdown of PointMetaBase~\cite{pointmetabase} on S3DIS~\cite{s3dis} and ScanNet~\cite{scannet}. The stacked bars show the proportion of time spent on FPS and other operations, demonstrating that FPS accounts for the dominant part of the total sampling latency.}
\label{fig:intro_overview}

\end{figure}

The FPS algorithm operates with a time complexity of $O(N \cdot M)$, where $N$ stands for the number of points in the point cloud, and $M$ stands for the number of target sampled points in the point cloud.
Selecting each of the $M$ samples requires a global distance comparison against all $N$ points, so a large volume of computation is spent merely to add a single point.
To quantify this impact, we profile FPS within PointMetaBase~\cite{pointmetabase} on a randomly selected 25\% subset of the validation samples from the indoor S3DIS~\cite{s3dis} and ScanNet~\cite{scannet} benchmarks (Fig.~\ref{fig:intro_overview}).
The top figure reports the point cloud size distribution together with the corresponding FPS sampling latency. Although most samples contain a moderate number of points, the FPS runtime increases sharply as the point cloud size grows. In particular, once the point cloud size exceeds $50\mathrm{k}$ points, the sampling cost becomes substantially higher, reaching around $250$ ms per sample and increasing to more than $1000$ ms for samples with over $100\mathrm{k}$ points. Therefore, the overall sampling cost is largely dominated by a small number of large-scale point clouds.
The bottom figure further shows the latency breakdown between FPS and the remaining operations. FPS accounts for the majority of the total latency on both datasets: $77.6\%$ on S3DIS and $89.0\%$ on ScanNet. This indicates that FPS is not merely an auxiliary preprocessing step, but the dominant computational bottleneck in the sampling pipeline. Moreover, ScanNet contains a higher proportion of large point clouds, especially those exceeding $100\mathrm{k}$ points, which explains its substantially higher average FPS latency and larger FPS ratio compared with S3DIS, even though the latency of the other operations is comparable.

Since FPS is inherently sequential, it maps poorly onto data-parallel GPUs, and most prior accelerators therefore rely on specialized hardware. FPGA-based designs such as MARS\cite{mars} and PtrAcc\cite{ptracc} exploit custom on-chip memory and data pipelines to achieve large speedups, but their hardware-specific optimizations do not transfer to general-purpose platforms. The KD-Tree-based QuickFPS\cite{quickfps} is portable to commodity hardware and is one of the strongest exact accelerators evaluated in this work, yet its substantial memory footprint is ill-suited to onboard robotic deployment. To accelerate FPS on processors without these limitations, we propose RadiusFPS, a spherical voxel-based algorithm whose radius pruning removes the redundant distance computations and memory traffic identified above while preserving the standard FPS update rule under a fixed seed and deterministic tie-breaking policy. To exploit GPU parallelism, we further introduce RadiusFPS-G, a warp-level implementation that fuses the core sampling sub-tasks into memory-coalesced kernels. Across indoor (S3DIS, ScanNet) and outdoor LiDAR (SemanticKITTI) benchmarks, our method advances efficiency on both hardware classes while maintaining competitive sampling quality. On the CPU, RadiusFPS accelerates sampling by up to 186× over vanilla FPS on large-scale scenes—far beyond the 6× of the approximate FPS+NPDU baseline. On the GPU, RadiusFPS-G reduces sampling latency by up to 52× relative to GPU FPS and consistently matches or exceeds QuickFPS among the evaluated methods while using only about half of its GPU memory. These gains carry through to full perception pipelines: as a drop-in module, RadiusFPS-G cuts End-to-End segmentation latency by up to 2.5×, and when paired with the learning-based FastPoint \cite{fastpoint} sampler, it raises the sampling speedup to 11.7× and the End-to-End speedup to 3.3× over the GPU-FPS baseline under the same backbone. Overall, RadiusFPS provides a strong exact-FPS-compatible accelerator among the evaluated CPU and GPU methods.

Our main contributions are listed as follows:
\begin{itemize}

    \item We identify redundant distance computations and excessive memory I/O as the principal sources of FPS inefficiency, and propose RadiusFPS, a spherical voxel-based sampling algorithm that addresses them through a dual-level pruning scheme: a radius-based voxel filter that discards entire irrelevant regions, followed by a coordinate-wise point-skip test that removes residual point updates. The spherical bound is provably conservative, so RadiusFPS accelerates sampling while preserving the FPS distance-update rule under the same initialization and tie-breaking policy.

    \item We further propose RadiusFPS-G, a GPU-accelerated variant that maps the inherently sequential sampling loop onto GPUs. Beyond conventional thread-level parallelism, it fuses voxel selection, radius pruning, and distance update into two warp-level fusion kernels with coalesced memory access, eliminating the global-memory round-trips that bottleneck existing GPU-based FPS implementations.

    \item  We establish a strong exact-FPS-compatible implementation across CPUs and GPUs among the evaluated methods: RadiusFPS reaches up to 186× over CPU FPS, while RadiusFPS-G matches or exceeds QuickFPS using only half of its GPU memory and comparable segmentation accuracy. Integrated into deep pipelines, both variants serve as drop-in modules and, combined with FastPoint, cut End-to-End latency by up to 3.3× over the GPU-FPS baseline, making FPS-style sampling practical for latency- and memory-constrained robotic perception.

\end{itemize}









\begin{figure}[t]
    \centering

    \includegraphics[width=0.65\linewidth]{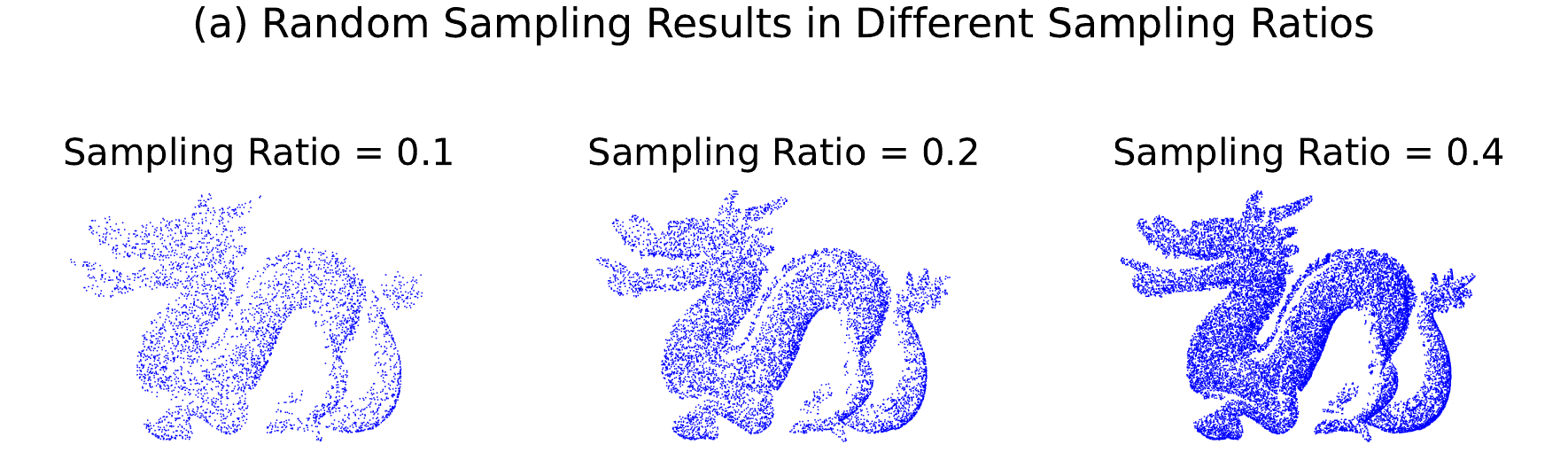}

    \vspace{0.4em}

    \includegraphics[width=0.65\linewidth]{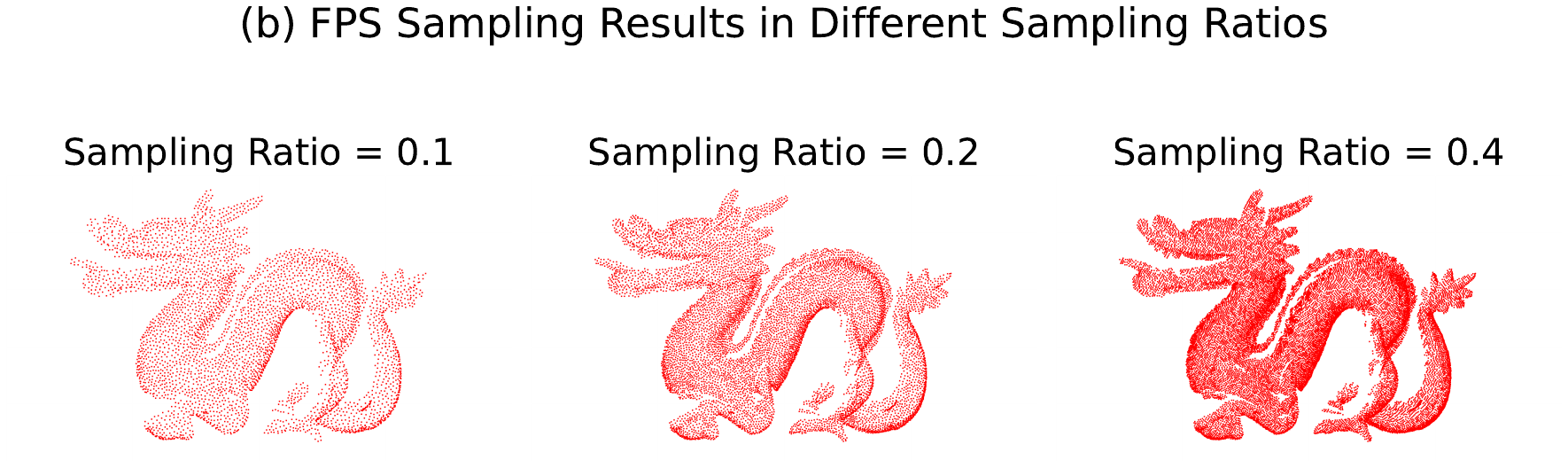}

    \vspace{0.4em}

    \includegraphics[width=0.65\linewidth]{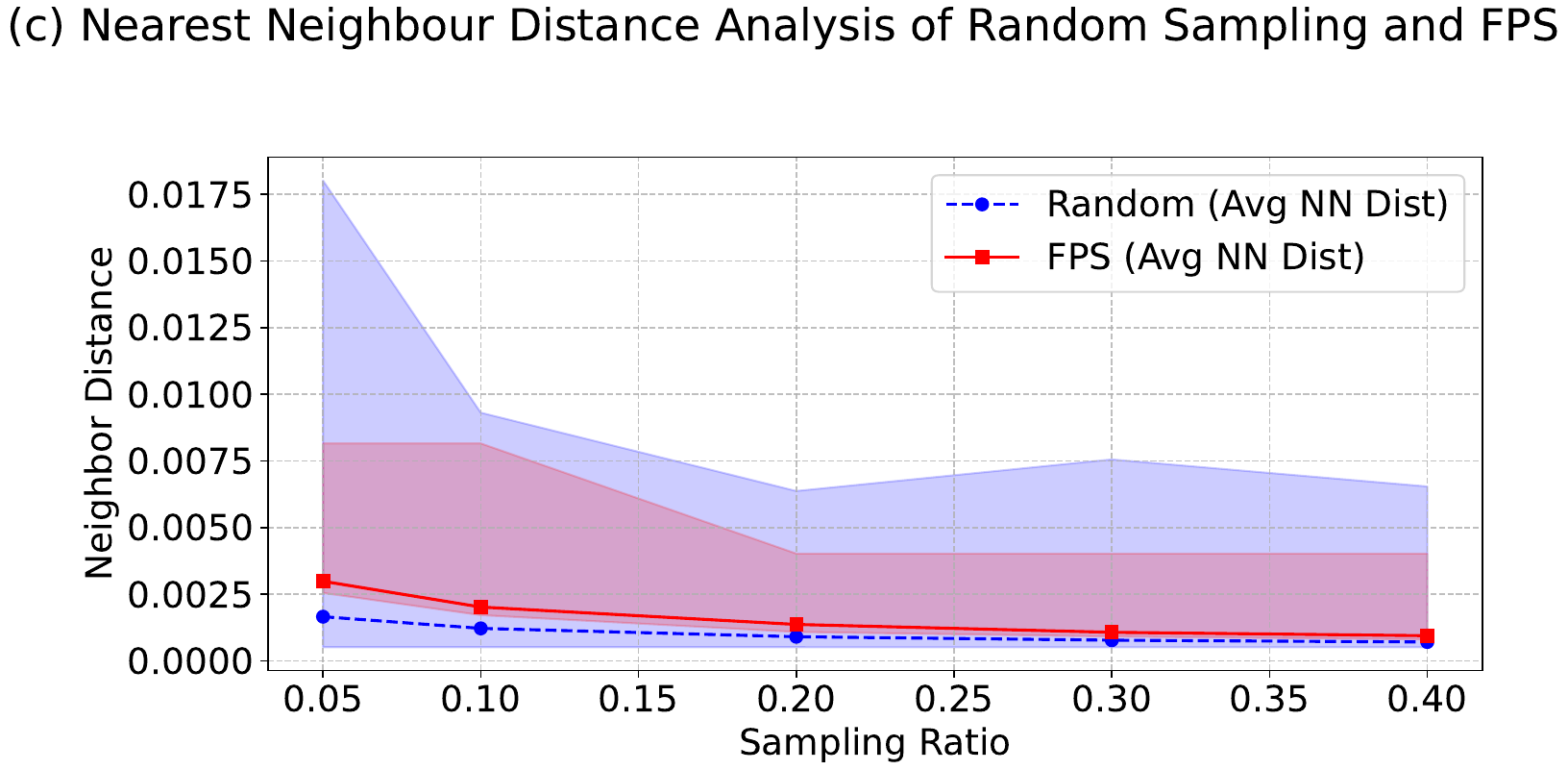}

    \caption{Sampling uniformity of FPS versus random sampling on the Stanford dragon point cloud. (a)-(b) sampling results at different ratios; (c) average nearest neighbor distance (lines) with min-max range(shaded).}
    \label{fig:sample_vis}
\end{figure}

\begin{algorithm}[t]
\caption{Standard FPS Algorithm}
\label{alg:fps}
\begin{algorithmic}[1]

\REQUIRE Input point cloud $P = \{p_1, p_2, \dots, p_N\}$; target sample size $M$
\ENSURE Sampled point set $S$

\STATE Initialize sampled set $S \leftarrow \emptyset$
\STATE Initialize distance array $D$ of size $N$ with $+\infty$
\STATE Randomly select a starting point $p_{start}$ from $P$
\STATE $S \leftarrow S \cup \{p_{start}\}$
\STATE $D[start] \leftarrow 0$

\WHILE{$|S| < M$}
    \STATE Let $s_{last}$ be the point most recently added to $S$
    \FOR{each point $p_i \in P$}
        \STATE $d \leftarrow \| p_i - s_{last} \|_2$
        \STATE $D[i] \leftarrow \min(D[i], d)$
    \ENDFOR
    \STATE $p_{next} \leftarrow \arg\max_{i} D[i]$
    \STATE $S \leftarrow S \cup \{p_{next}\}$
\ENDWHILE

\RETURN $S$

\end{algorithmic}
\end{algorithm}

\section{Background Knowledge}
\subsection{Farthest Point Sampling}
Initially popularized by PointNet++~\cite{pointnet++}, Farthest Point Sampling (FPS) has become one of the most standard sampling algorithms for point cloud-based deep learning models~\cite{fps-choy20194d,fps-qi2018frustum,pointbert,pointmae,pointmetabase,pointvector,pointrcnn,pointransformer}. Beyond simply downsampling dense point clouds, FPS effectively preserves representative points that capture local structural information, thereby facilitating subsequent processing tasks. Numerous studies have since validated the efficacy of FPS in deep learning applications~\cite{3dssd,IASSD,qiu2021semantic}.

The detailed procedure of the standard FPS is formalized in Alg.~\ref{alg:fps}. Given an input point cloud $P = \{p_1, p_2, \dots, p_N\}$ and a target sample size $M$, the algorithm maintains a distance array $D$ of size $N$ to track the shortest distance from each point to the current sampled set $S$. Initially, all entries of the distance array are set to $+\infty$, i.e., $D[i] = +\infty$ for all $i = 1, \dots, N$, and the sampled set $S$ is seeded with a randomly selected starting point $p_{start}$. In each iteration of sampling, rather than recomputing distances to the entire set $S$, we efficiently update $D$ based on the most recently added point, denoted as $s_{last}$. For every point $p_i \in P$, we compute the Euclidean distance to $s_{last}$ and update the stored distance only if the new value is smaller (i.e., $D[i] \leftarrow \min(D[i], \|p_i - s_{last}\|_2)$). This ensures that $D[i]$ always represents the minimum distance from $p_i$ to any point in $S$. Finally, the point with the maximum value in $D$—representing the point most distant from the existing samples—is selected as $p_{next}$ ($p_{next} \leftarrow \text{argmax}(D)$) and added to $S$ for the next cycle. The algorithm then proceeds iteratively until the target size $M$ is reached.

A primary advantage of FPS is its inherent ability to regulate spatial distribution, ensuring that the sampled subset provides a comprehensive representation of the underlying geometry.
Fig.\ref{fig:sample_vis} shows the sampling results on the dragon point cloud from the Stanford 3D Scanning Dataset~\cite{stanfordscan} and presents a quantitative analysis of neighbor distances.

As shown in the visual comparison between Fig.~\ref{fig:sample_vis}(a) and Fig.~\ref{fig:sample_vis}(b), random sampling exhibits a highly unbalanced distribution characterized by dense clusters and significant voids.

In contrast, FPS yields a much more homogeneous pattern, effectively preserving structural details—such as edges and surface contours—across all sampling ratios.
Fig.~\ref{fig:sample_vis}(c) further quantifies this stability. The line chart tracks the average Nearest Neighbor (NN) distance, while the shaded regions (blue for Random, red for FPS) represent the range (spread between maximum and minimum) of these distances. Notably, the extensive blue region completely encompasses the narrower red region. This stark contrast indicates that while Random Sampling suffers from extreme irregularities—points being either overly clustered or sparsely scattered—FPS maintains a consistent inter-point distance. This "tight" distribution control ensures that local features are uniformly captured without redundancy or main information loss.

\subsection{Challenges of FPS}
Given the above introduction, we can conclude that the challenges in accelerating the FPS algorithm fall into three main aspects:
\begin{itemize}
    \item \textbf{High computational overhead.}
    FPS necessitates exhaustive Euclidean distance calculations and comparisons for the entire point cloud in each round. As the number of input points ($N$) and target samples ($M$) increases, the computational complexity grows significantly (typically $O(N \cdot M)$), becoming a bottleneck for large-scale processing.


    \item \textbf{Lack of Filtering Mechanism.}
    In each iteration, the standard FPS performs a global distance update over all points:
    \begin{equation}
        \mathrm{Dist}_p[i] \leftarrow \min\!\big(\mathrm{Dist}_p[i],\ \lVert p_i - s_{\text{last}}\rVert_2\big),
        \quad i = 0,\ldots,N-1 ,
    \end{equation}
    where $\mathrm{Dist}_p[i]$ is the current nearest sample distance of $p_i$ and $s_{\text{last}}$ is the most recently selected sample. However, only points satisfying $\lVert p_i - s_{\text{last}}\rVert_2 < \mathrm{Dist}_p[i]$ actually change; the rest remain unchanged. Standard FPS lacks any mechanism to identify and skip these unaffected points, causing significant computational redundancy.

    \item \textbf{Inherent Sequential Dependency.}
    Since each new sample is selected based on the distances to the previously sampled set, there exists a strict dependency between iterations. This iterative nature prevents the simultaneous acquisition of all points via simple parallelization techniques.
    The sampling process is sensitive to local deviations; an erroneous selection of a single point can propagate errors to subsequent iterations, potentially compromising the overall quality of the sampled set.

\end{itemize}

\begin{figure}
    \centering
    \includegraphics[width=0.6\linewidth]{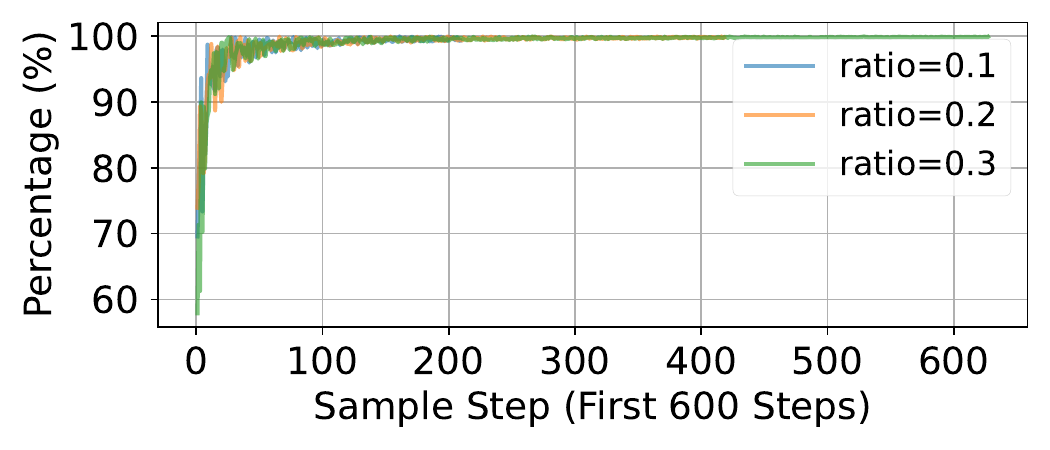}
    \caption{Percentage of ineffective distance updates at each sampling step of standard FPS on the Stanford dragon point cloud. Across different sampling ratios, ineffective updates rapidly dominate after the initial stage and remain above 90\% in most later iterations, indicating that most distance computations and memory accesses do not change the stored nearest-sample distances.}
    \label{fig:samplestep}
\end{figure}




Among the challenges above, the lack of a filtering mechanism is the most critical for practical acceleration. In each FPS iteration, the newly selected sample is compared with all (N) input points, although only a small subset may actually reduce their stored nearest sample distances. We define a distance update as \emph{effective} if
\begin{equation}
    \lVert p_i - s_{\mathrm{last}}\rVert_2 < Dist_p[i],
\end{equation}

and \emph{ineffective} otherwise. Ineffective updates leave $Dist_p[i]$ unchanged and therefore do not affect subsequent sample selection, while still incurring distance computation and memory-access costs.

To quantify this redundancy, we profile standard FPS on the Stanford dragon point cloud under three sampling ratios. As shown in Fig.~\ref{fig:samplestep}, ineffective updates rapidly dominate after the initial sampling stage. The first several iterations still cover large unexplored regions, but later samples typically influence only a local neighborhood because most points already have nearby representatives. Consequently, more than 90\% of distance computations become ineffective in later iterations. This observation motivates the coarse-to-fine pruning strategy of RadiusFPS, which first eliminates unaffected spatial regions and then skips unnecessary point-level updates.

\subsection{Related Works}
Existing efforts to accelerate FPS fall into two broad categories: algorithm structure optimization and sampling pattern optimization. The former preserves the exact FPS output and reduces its cost by redesigning the data structures, memory access, or underlying hardware on which sampling runs; the latter relaxes exactness, substituting parallel-friendly heuristics or learned predictors that sample faster but no longer reproduce the FPS result. We review each category below and position our method relative to both.

\subsubsection{Algorithm Structure Optimization}
A range of accelerators speed up FPS by redesigning its data structures or the hardware on which it runs, while keeping the sampled output exact~\cite{pointacc,ptracc,mars,ptracc++,quickfps,PSDfps}. On general-purpose hardware, QuickFPS\cite{quickfps} is the strongest representative: Han et al. organize the point cloud with a KD-tree to prune distance computations and regularize memory access, making it the fastest existing exact accelerator. This speed, however, comes at the cost of a substantial memory footprint that is unsuited to memory-constrained onboard platforms. A second line of work turns to custom silicon: MARS\cite{mars} and PtrAcc\cite{ptracc} build FPGA-based pipelines that cut distance computation and streamline the sampling workflow, achieving large speedups without sacrificing accuracy. Their gains, however, hinge on FPGA-specific on-chip RAM and dataflow and do not transfer to general-purpose CPUs or GPUs. In short, structure-level methods preserve exact FPS results but remain either memory-heavy or hardware-bound.

\subsubsection{Sampling Pattern Optimization}
To bypass the sequential cost of FPS, a second line of work replaces it with parallel-friendly sampling patterns. Grid sampling\cite{pcl} is the most common alternative: it partitions the bounding box of the point cloud into regular cells and keeps, in each cell, the point closest to the cell center. This gives better control over the average inter-point spacing and has been adopted by models such as Grid-GCN\cite{gridgcn}, KPConv\cite{kpconv}, and PGFormer\cite{pgformer}. Other methods instead approximate the FPS pattern to reduce the number of sampling rounds. Targeting on-device perception, EdgePC\cite{edgepc} encodes the point cloud with Morton codes to enable a fully parallel sampling strategy for edge platforms, while NPDU\cite{afps} segments a LiDAR scan and selects multiple points per round via an efficient nearest neighbor search. By committing several points in a single iteration, these heuristics save substantial processing time but fall short of FPS in sampling quality, which in turn degrades downstream task performance.

A more recent direction learns the sampling pattern from FPS itself. FastPoint\cite{fastpoint} analyzes the sampling positions and distribution produced by the first few FPS iterations and predicts a sampling curve, thereby avoiding the full iterative FPS process with only minor quality degradation. However, because it still bootstraps from standard FPS to generate the initial trajectory and relies on a learned prediction module, FastPoint is regarded as an FPS-based acceleration component for learning-based pipelines rather than a standalone sampling strategy.


\begin{figure*}[t]
\centering

\subfloat[Point Cloud Preprocessing]{
    \includegraphics[width=0.5\textwidth]{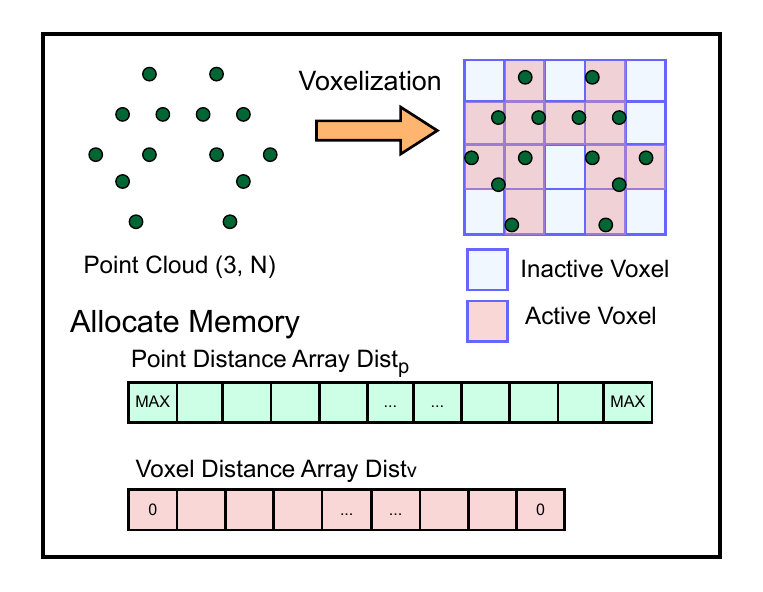}
    \label{fig:radiusfps_a}
}
\subfloat[ First Point Selection and Distance Initialization]{
    \includegraphics[width=0.5\textwidth]{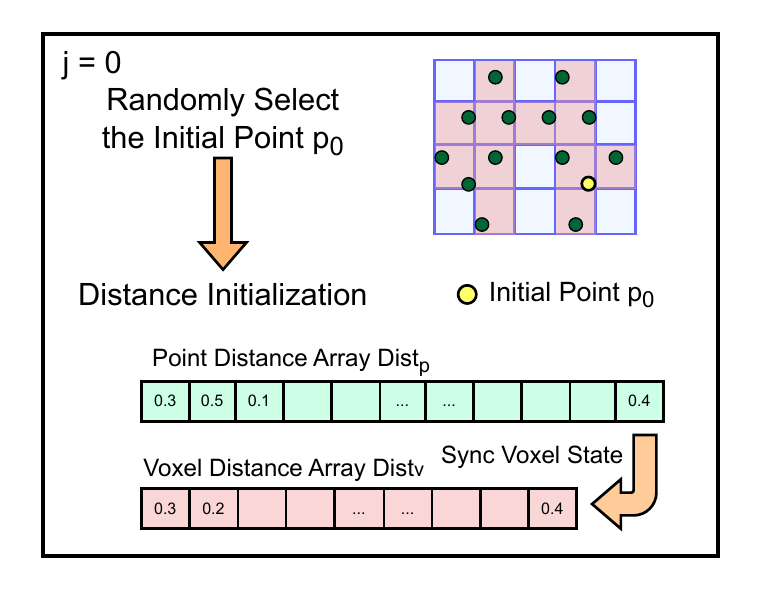}
    \label{fig:radiusfps_b}
}

\vspace{0.3em}

\subfloat[Iterative Sampling with Voxel-Level Pruning and Point Skip]{
    \includegraphics[width=\textwidth]{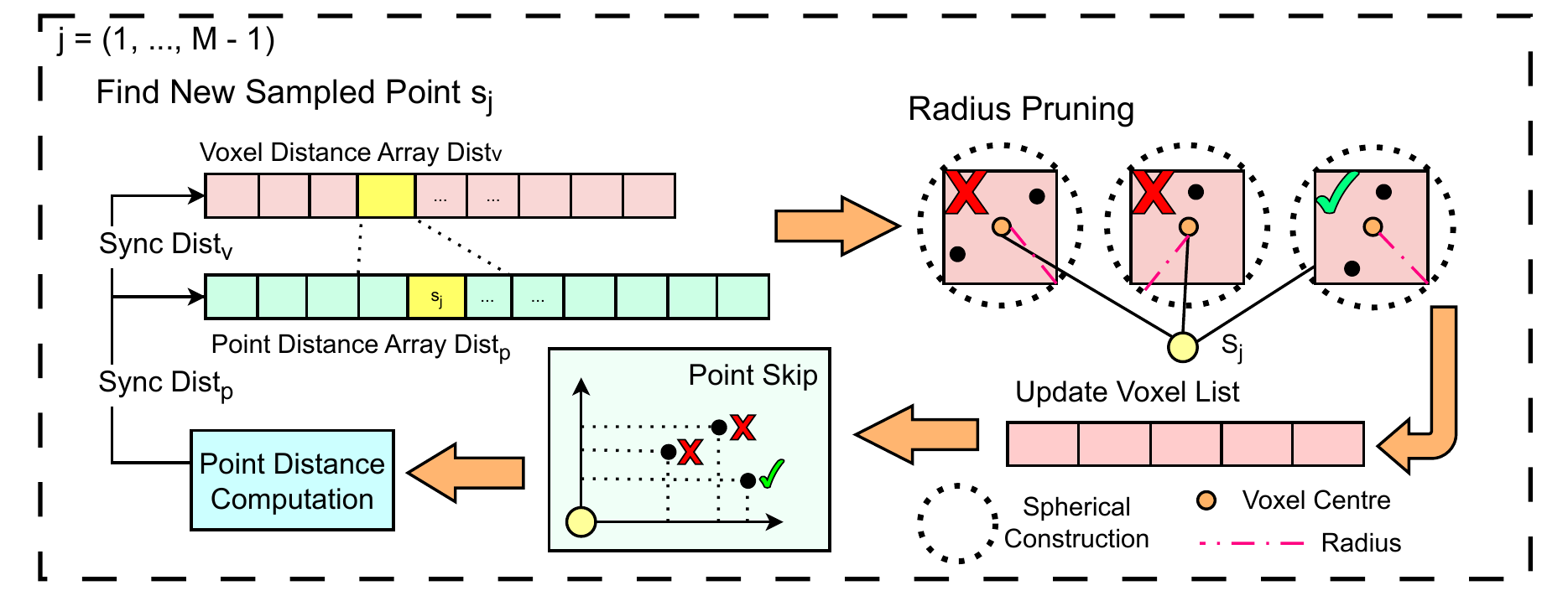}
    \label{fig:radiusfps_c}
}

\caption{An Overview of the RadiusFPS Algorithm. The algorithm mainly consists of three stages: (a) Point Cloud Preprocessing: Voxelize the input point cloud and allocate memory for voxel and point distance array $Dist_p, Dist_v$. (b) Initial update: Randomly select the initial sampled point $p_0$, update distance arrays with global Euclidean distance computation. (c) Iterative sampling with voxel-level pruning and point skip: The sampling process can be split into three steps: firstly, find the new sampled point, then utilize efficient radius pruning and point skip strategy to decide a subset of points for distance update. Eventually, only selected points are required to compute an accurate distance and perform memory access.}
\label{fig:radiusfps}

\end{figure*}

\section{Method}
\label{sec:method}

In this section, we present RadiusFPS, a spherical-voxel-based acceleration framework for exact farthest point sampling. As illustrated in Fig.~\ref{fig:radiusfps}, RadiusFPS follows three main stages. First, the input point cloud is partitioned into active voxels, and two distance states are maintained: a point-level nearest sample distance array ($Dist_p$) and a voxel-level maximum-distance array ($Dist_v$) (Fig.\ref{fig:radiusfps_a}). Second, RadiusFPS selects an initial seed point ($p_0$) and performs a global distance initialization. Specifically, ($Dist_p$) is initialized by computing the distance from each point to ($p_0$), while ($Dist_v$) is obtained by taking the maximum ($Dist_p$) value within each active voxel (Fig.~\ref{fig:radiusfps_b}). Finally, the algorithm enters the iterative sampling stage. At each iteration, the next sampled point is first selected from the current distance states. RadiusFPS then applies conservative voxel-level radius pruning to discard voxels that cannot contain effective updates, followed by a coordinate-wise point-skip test to avoid unnecessary point-level distance computations inside the remaining voxels (Fig.~\ref{fig:radiusfps_c}). This coarse-to-fine pruning strategy reduces redundant distance computations and memory accesses while preserving the update behavior of standard FPS. The following subsections describe each stage in detail.

\subsection{Notation}
Given an input point cloud $P = \{p_0, p_1, \dots, p_{N-1}\}$ with $N$ points, where $p_i \in \mathbb{R}^3$, the goal of FPS is to select a subset $S = \{s_0, s_1, \dots, s_{M-1}\}$ of $M$ points that provides representative coverage of the original set. The sampling process is performed iteratively. Let $S_{j-1} = \{s_0, s_1, \dots, s_{j-1}\}$ denote the set of points selected before the $j$-th selection. The next point $s_j$ is chosen as the point in $P$ whose nearest distance to the current sampled set is maximized:
\begin{equation}
s_j = \arg\max_{p \in P} \left( \min_{s \in S_{j-1}} \|p - s\|_2 \right).
\end{equation}
To avoid recomputing the distance from every point to all previously selected samples at each iteration, FPS maintains a point-level distance array $Dist_p$. Each element $Dist_p[i]$ records the current nearest sample distance of point $p_i$, namely
\begin{equation}
Dist_p[i] = \min_{s \in S} \|p_i - s\|_2.
\end{equation}
With this array, each iteration only needs to compute the distance from all points to the newly selected sample and update $Dist_p$ when a smaller distance is found. To make exactness well-defined, all FPS variants in our algorithmic discussion use the same initial seed, the same floating-point precision, and deterministic tie-breaking: when multiple candidates have the same distance, $\arg\max$ returns the point with the smallest original input index. Different seeds or tie-breaking policies can yield different valid FPS sequences and slightly different downstream metrics.
We further partition \(P\) into \(K\) active voxels
\(\mathcal{V}=\{v_0,v_1,\ldots,v_{K-1}\}\). For each active voxel \(v_k\), RadiusFPS maintains a voxel-level distance value
\begin{equation}
    Dist_v[k]=\max_{p_i\in v_k} Dist_p[i],
\end{equation}

which records the largest current nearest sample distance among points in \(v_k\). This value serves as the voxel-level threshold used by the pruning strategy introduced below.


\subsection{Preprocessing}
\label{sec:preprocessing}

In the preprocessing stage, RadiusFPS voxelizes the input point cloud to establish a compact spatial index for subsequent voxel-level pruning. Let the axis-aligned bounding box of the input point cloud be defined by $p_{min} =(x_{min}, y_{min}, z_{min}), p_{max} = (x_{max}, y_{max}, z_{max})$ and let its side lengths be
\begin{equation}
    E = (E_x, E_y, E_z) = (x_{max} - x_{min}, y_{max} - y_{min}, z_{max} - z_{min})
\end{equation}

To ensure all points $p \in P$ are contained within the voxel grid, the bounding box is slightly scaled by an expansion factor $\alpha = 1 + \varepsilon$ (where $\varepsilon \ll 1$). The side length of a single voxel $L$ is then determined by a user-defined voxel resolution $n_{\mathrm{vox}}$, which denotes the number of voxel bins along each axis rather than a physical distance (reported as $v$ in the parameter-sensitivity experiments):
\begin{equation}
L = \frac{\alpha \max\{E_x,E_y,E_z\}}{n_{\mathrm{vox}}}.
\label{eq:voxel_size}
\end{equation}
For every point $p_i = (x_i, y_i, z_i) \in P$, its position is mapped to discrete integer coordinates $(v_x, v_y, v_z)$ representing its home voxel. Boundary values are clamped to keep every coordinate in the valid range:
\begin{equation}
v_{i,a}=\min\!\left(n_{\mathrm{vox}}-1,\max\!\left(0,\left\lfloor \frac{p_i^a-p_{min}^a}{L}\right\rfloor\right)\right),
\quad a\in\{x,y,z\}.
\end{equation}


To handle the inherent sparsity of point clouds, RadiusFPS stores only active voxels, namely, voxels that contain at least one input point. Empty voxels are discarded and never visited during sampling. Let the set of active voxels be denoted by
\begin{equation}
\mathcal{V}={v_0,v_1,\ldots,v_{K-1}},
\end{equation}
where $K$ is the number of active voxels. For each active voxel $v_k$, RadiusFPS records the indices of its contained points and maintains its voxel center $c_k$. In addition, each voxel is associated with a bounding sphere radius
\begin{equation}
r_k=\frac{\sqrt{3}}{2}L,
\label{eq:preprocess_voxel_radius}
\end{equation}
which guarantees that all points inside $v_k$ are enclosed by the sphere centered at $c_k$. This spherical representation is later used to derive a conservative lower bound for voxel-level pruning.

After the active voxel index is constructed, RadiusFPS allocates two distance-tracking structures. The point-level distance array $Dist_p\in\mathbb{R}^{N}$ stores the current nearest sample distance of each point and is initialized to $+\infty$. The voxel-level distance array $Dist_v\in\mathbb{R}^{K}$ stores the maximum current point-level distance within each active voxel:
\begin{equation}
Dist_v[k]=\max_{p_i\in v_k} Dist_p[i].
\label{eq:preprocess_distv}
\end{equation}
Before the first sample is selected, $Dist_v$ is initialized to zero and is synchronized with $Dist_p$ after the initial global distance update. Therefore, this preprocessing stage only builds the sparse voxel index and prepares the distance states, while the actual distance values are established in the initialization stage described next.


\subsection{Initial Update}
The initial iteration ($j=0$) serves as the priming phase for the RadiusFPS algorithm. While subsequent iterations leverage hierarchical pruning to bypass redundant regions, this step is dedicated to establishing the global distance state required to initialize the two-level acceleration.

\textbf{Seed Selection. }
The process begins by selecting the primary seed point $s_0$ from the input point cloud $P$. To ensure an unbiased starting state, $s_0$ is typically chosen via a uniform discrete distribution over the indices of the point set (as illustrated in Fig.\ref{fig:radiusfps_b}):
\begin{equation}
    s_0 = p_k, \quad k \sim \text{Uniform}(0, N-1)
\end{equation}

\textbf{Distance Initialization}
Since no prior distance information exists to facilitate pruning at this stage, the algorithm performs an exhaustive global update. For every point $p_i \in P$, the algorithm computes the exact Euclidean distance to the initial seed point $s_0$ and initializes the point-level distance array as:
\begin{equation}
    Dist_p[i] = \| p_i - s_0 \|_2, \quad \forall i \in \{0, \dots, N-1\}
\end{equation}
This step incurs a computational cost of $O(N)$, ensuring that the foundation for subsequent sampling is mathematically rigorous. After the point-level distances are initialized, RadiusFPS synchronizes the voxel-level distance array by taking the maximum point-level distance inside each active voxel:
\begin{equation}
    Dist_v[k] =\max_{p_i \in v_k} Dist_p[i]
\end{equation}
Consequently, $Dist_p$ records the nearest sample distance of each point to the current sampled set $S={s_0}$, while $Dist_v[k]$ records the largest such distance among points in voxel $v_k$. These two synchronized distance states form the basis for the hierarchical sample selection and pruning operations in the subsequent iterations.


\subsection{Efficient Iterative Sampling}

\subsubsection{Find New Sample Point}

Following the initial update, the algorithm proceeds to the main loop to iteratively select the remaining $M-1$ samples. Given that the initial point-level and voxel-level distance states have been established, each iteration begins by identifying the next farthest sample point.

We reformulate the selection of the $j$-th sampled point $s_j$ as a two-level maximization problem. Instead of scanning all points in $P$ directly, RadiusFPS first searches over the active voxels and then searches within the selected voxel. Specifically, the index of candidate voxel $k^{*}$ is determined by
\begin{equation}
k^{*}=\arg\max_{0\leq k<K} Dist_v[k],
\label{eq:best_voxel}
\end{equation}
where $Dist_v[k]$ denotes the maximum current point-level distance among all points inside voxel $v_k$. Then, the next sampled point is selected from this candidate voxel:
\begin{equation}
s_j=\arg\max_{p_i\in v_{k^{*}}} Dist_p[i],
\quad j\in{1,\dots,M-1}.
\label{eq:best_point}
\end{equation}
Since ($Dist_v[k]$) is synchronized as
\begin{equation}
Dist_v[k]=\max_{p_i\in v_k}Dist_p[i],
\label{eq:distv_sync_selection}
\end{equation}
the above two-level selection is equivalent to the global FPS selection rule when the same deterministic tie-breaking policy is used at both the voxel and point levels:
\begin{equation}
s_j=\arg\max_{p_i\in P}Dist_p[i].
\label{eq:global_equivalent_selection}
\end{equation}

Therefore, the hierarchical search changes only how the farthest point is located, while preserving the sample-selection behavior of standard FPS under the fixed seed and tie-breaking policy stated above. After $s_j$ is selected, RadiusFPS updates the distance states using the voxel-level radius pruning and point-level skip strategies described in the following subsections.

\begin{figure}[t]
    \centering
    \includegraphics[width=0.7\linewidth]{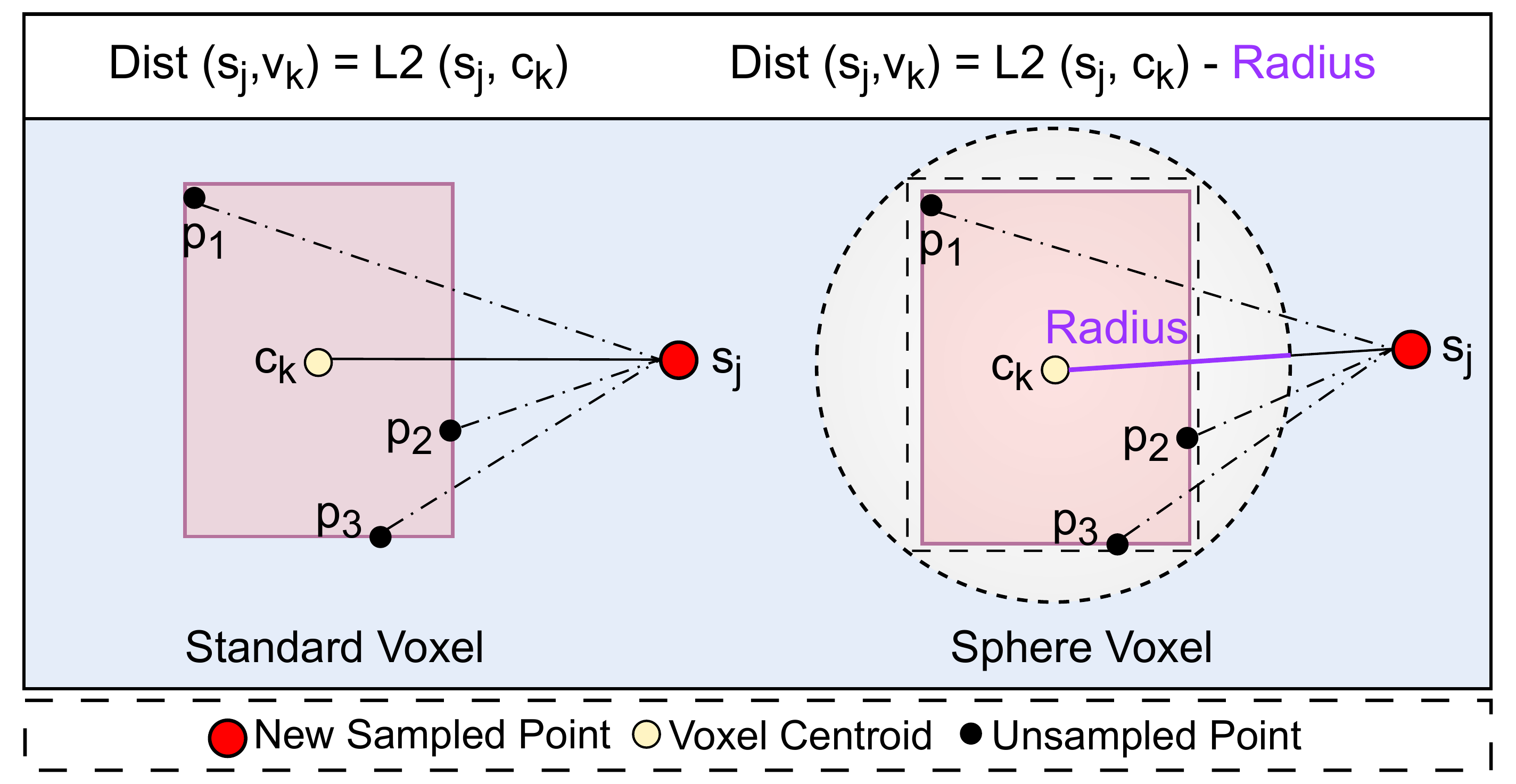}
    \caption{Standard Voxel Design (Left) and Spherical Voxel Design (Right). The main difference between the two voxels is the sphere construction covering all the points inside the voxel. In this figure, for the new sampled point $s_j$, the distance between $s_j$ and the edge of the spherical voxel provides a conservative lower bound on point distances to $s_j$. For the standard voxel, on the contrary, a similar centroid-only strategy may lead to incorrect sampling in the next iteration.}
    \label{fig:voxeldesign}
\end{figure}

\begin{figure*}[t]
    \centering
    \includegraphics[width=\linewidth]{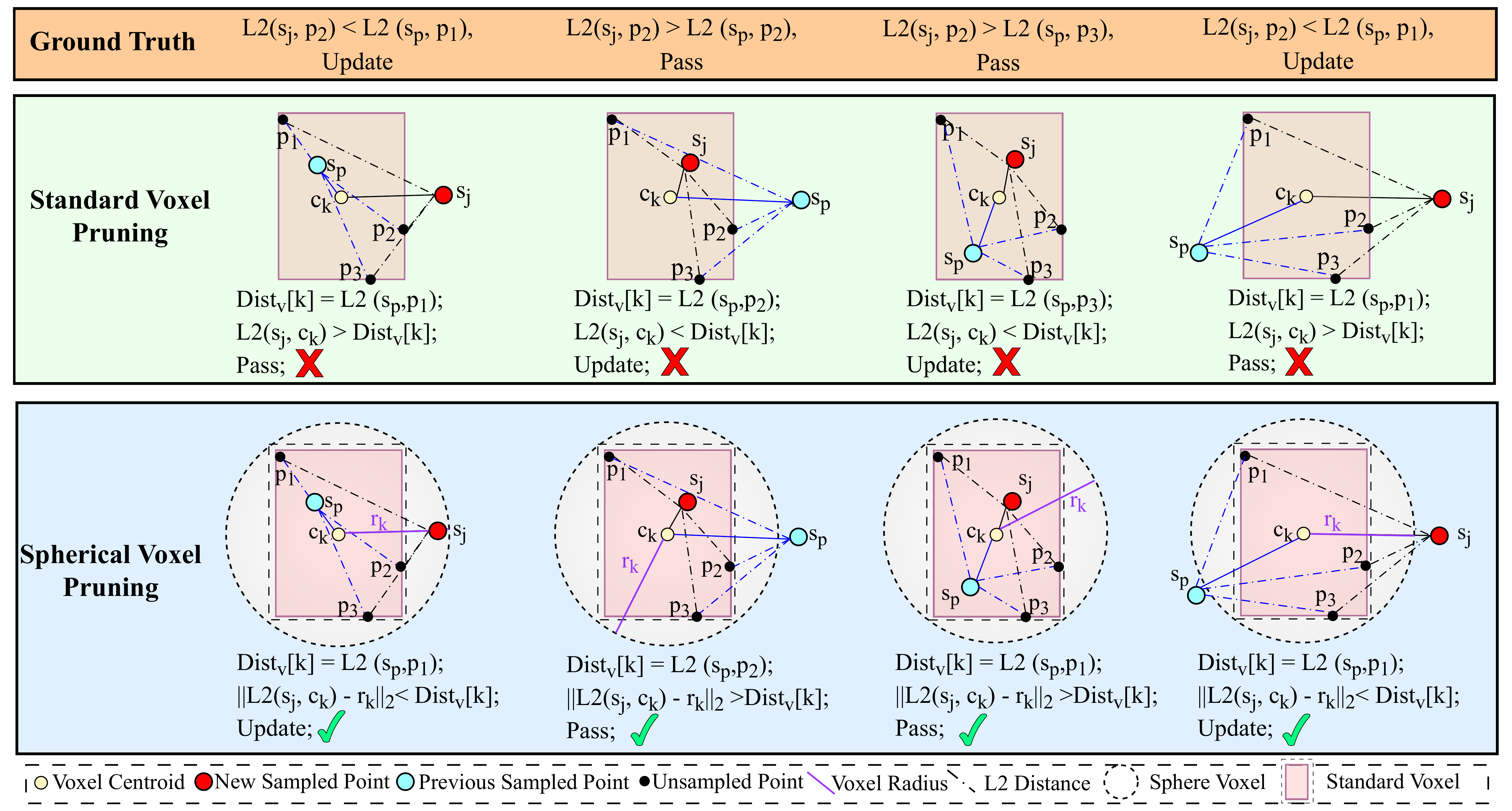}
    \caption{Illustration of the standard voxel pruning versus our proposed spherical voxel pruning. The top row defines the ground truth for updating or passing based on exact Euclidean distances. The standard voxel pruning (middle) fails in the illustrated cases (marked with red crosses) because relying solely on the distance to the voxel centroid $c_k$ provides an unsafe geometric criterion. By incorporating a bounding sphere with radius $r_k$, our spherical voxel pruning (bottom) uses the conservative lower bound $\max(0,\|s_j-c_k\|_2-r_k)$ and avoids unsafe false pruning in these cases (green checkmarks).}
    \label{fig:pruncases}
\end{figure*}

\subsubsection{Radius Pruning Based on Spherical Voxels}
\label{sec:radiuspruning}


In standard FPS, each iteration updates $Dist_p$ by comparing the newly selected sample $s_j$ with every point in $P$. Although this exhaustive update is exact, most comparisons do not reduce the stored nearest sample distance and therefore have no effect on subsequent sample selection. The resulting redundant distance computations and memory accesses constitute the main opportunity for acceleration.

RadiusFPS addresses this redundancy by first deciding whether an entire voxel can be safely skipped before accessing its internal points. A straightforward approach is to represent each voxel $v_k$ by its centroid $c_k$ and use the center distance $\|s_j-c_k\|_2$ as an estimate of the distance from $s_j$ to all points in the voxel. However, such a centroid-based criterion is not safe for exact FPS. As illustrated in Fig.~\ref{fig:voxeldesign}, the center distance is not necessarily a lower bound for all point-wise distances inside the voxel. In particular, there may exist a point $p_i\in v_k$, such as $p_2$ in the figure, satisfying
\begin{equation}
\|s_j-p_i\|_2 < \|s_j-c_k\|_2 .
\end{equation}
In this case, pruning the voxel according only to $\|s_j-c_k\|_2$ may incorrectly skip a point whose stored distance $Dist_p[i]$ should be updated. Since FPS selects each subsequent sample from the updated distance array, a single unsafe skip may change the following sampling sequence.




Therefore, a safe voxel-level pruning rule must rely on a conservative lower bound rather than on a centroid-only distance estimate. Specifically, for each voxel $v_k$, we require a quantity $LB(v_k,s_j)$ satisfying
\begin{equation}
LB(v_k,s_j) \leq \|s_j-p_i\|_2,
\quad \forall p_i\in v_k .
\label{eq:voxel_lower_bound_requirement}
\end{equation}
If this lower bound is no smaller than the maximum current distance stored in the voxel, then no point in the voxel can be updated by $s_j$.

To obtain such a bound, RadiusFPS represents each active voxel $v_k$ by a bounding sphere centered at $c_k$, as shown in Fig.~\ref{fig:voxeldesign}. Since all points in $v_k$ are enclosed by this sphere, the sphere radius is determined by the voxel side length $L$:
\begin{equation}
r_k=\frac{\sqrt{3}}{2}L .
\label{eq:spherical_voxel_radius}
\end{equation}
For any point $p_i\in v_k$, we have $\|p_i-c_k\|_2\leq r_k$. According to the triangle inequality,
\begin{equation}
\|s_j-p_i\|_2
\geq
\|s_j-c_k\|_2-\|p_i-c_k\|_2 .
\label{eq:triangle_inequality_bound}
\end{equation}
Combining Eq.~\eqref{eq:triangle_inequality_bound} with $\|p_i-c_k\|_2\leq r_k$, RadiusFPS defines the lower bound from $s_j$ to voxel $v_k$ as
\begin{equation}
LB(v_k,s_j)
=
\max\left(0,\|s_j-c_k\|_2-r_k\right).
\label{eq:spherical_voxel_bound}
\end{equation}
By construction, this bound satisfies the requirement in Eq.~\eqref{eq:voxel_lower_bound_requirement}:
\begin{equation}
LB(v_k,s_j)\leq \|s_j-p_i\|_2,
\quad \forall p_i\in v_k.
\label{eq:lower_bound_property}
\end{equation}

Based on this lower bound, RadiusFPS skips voxel $v_k$ when
\begin{equation}
LB(v_k,s_j) \geq Dist_v[k],
\label{eq:radius_pruning_condition}
\end{equation}
where $Dist_v[k]=\max_{p_i\in v_k}Dist_p[i]$ denotes the largest current nearest sample distance among points in $v_k$. This condition is safe because, for any point $p_i\in v_k$, we have
\begin{equation}
\|s_j-p_i\|_2
\geq LB(v_k,s_j)
\geq Dist_v[k]
\geq Dist_p[i].
\label{eq:radius_pruning_correctness}
\end{equation}
Therefore, no point inside $v_k$ can satisfy $\|s_j-p_i\|_2<Dist_p[i]$, which means that all point-level distance records in this voxel remain unchanged. The equality case is also safe because the FPS distance update only changes $Dist_p[i]$ when a strictly smaller distance is found. Consequently, pruning $v_k$ does not alter the distance-update behavior or the subsequent sample selection of standard FPS.

If the pruning condition is not satisfied, voxel $v_k$ may still contain points whose distance records can be reduced. RadiusFPS then visits the points inside $v_k$ and applies the coordinate-wise point-skip test introduced in the next subsection before performing exact Euclidean distance computation.


Fig.~\ref{fig:pruncases} further illustrates why the spherical bound is necessary. In these representative cases, the current distance records in voxel $v_k$ are determined by a previously selected sample $s_p$, and the algorithm must decide whether the newly selected sample $s_j$ can reduce any point-level distance in the voxel. The positions of $s_p$ and $s_j$ cover both inside-voxel and outside-voxel configurations.

Centroid-based pruning can make incorrect decisions because the centroid does not constrain the minimum possible distance from $s_j$ to every point in $v_k$. In contrast, the spherical representation explicitly encloses all points in the voxel and therefore provides a valid lower bound for every point-wise distance. As a result, spherical voxel pruning safely identifies voxels whose point distances cannot be updated; voxels that fail the pruning test are conservatively retained for the point-level skip and exact-distance update stages.

\subsubsection{Point Skip Acceleration}
Radius pruning removes entire voxels whose point-level distances are guaranteed to remain unchanged. However, a voxel that survives Eq.~\eqref{eq:radius_pruning_condition} does not imply that every point inside it must be updated. In many cases, only a small subset of points in the surviving voxel can potentially obtain a smaller nearest sample distance from the newly selected sample $s_j$. Therefore, RadiusFPS further applies a point-level skip test before performing exact Euclidean distance computation.

For each point $p_i$ in a visited voxel, an exact update is needed only when the new distance to $s_j$ is smaller than the current record $Dist_p[i]$. RadiusFPS uses the relation between the Euclidean norm and the $L_\infty$ norm:
\begin{equation}
    \| p_i - s_j \|_2
    \geq
    \|p_i - s_j\|_\infty
    = \max_{a\in\{x,y,z\}} |p_i^a-s_j^a| .
    \label{eq:point_skip_linf_bound}
\end{equation}
Accordingly, if the coordinate difference along any axis is already no smaller than the current nearest sample distance, then the exact Euclidean distance cannot reduce $Dist_p[i]$:
\begin{equation}
    \text{skip } p_i
    \quad \text{if} \quad
    \exists a\in\{x,y,z\}: |p_i^a-s_j^a| \geq Dist_p[i].
    \label{eq:point_skip_condition}
\end{equation}
This condition is conservative. It may leave some non-updatable points for exact computation, but it never skips a point whose distance record should be decreased. Therefore, the point-skip test preserves the exact update behavior of FPS while avoiding a large number of unnecessary square, summation, and square-root operations.

Only points that pass Eq.~\eqref{eq:point_skip_condition} are subjected to the exact distance update:
\begin{equation}
    Dist_p[i] \leftarrow
    \min \left(
    Dist_p[i],
    \sqrt{\sum_{a \in \{x,y,z\}} (p_i^{a} - s_j^{a})^2}
    \right).
    \label{eq:point_distance_update}
\end{equation}
From a geometric perspective, Eq.~\eqref{eq:point_skip_condition} characterizes a necessary condition for exact distance computation: $p_i$ must lie inside the axis-aligned cube centered at $s_j$ with half side length $Dist_p[i]$. Points outside this cube are guaranteed to be farther than their current nearest sample distance and can be skipped safely.

After all necessary point-level updates in a visited voxel are completed, RadiusFPS synchronizes the voxel state by recomputing the maximum residual distance within that voxel:
\begin{equation}
    Dist_v[k] \leftarrow \max_{p_i \in v_k} Dist_p[i].
    \label{eq:point_skip_distv_sync}
\end{equation}
For voxels pruned by Eq.~\eqref{eq:radius_pruning_condition}, both $Dist_p$ and $Dist_v$ remain unchanged and no synchronization is required. Together, spherical radius pruning and point skip form a coarse-to-fine filtering strategy: the former avoids accessing irrelevant voxels, whereas the latter avoids exact distance computation for redundant points inside the remaining voxels.

\begin{figure*}[h]
    \centering
    \includegraphics[width=\textwidth]{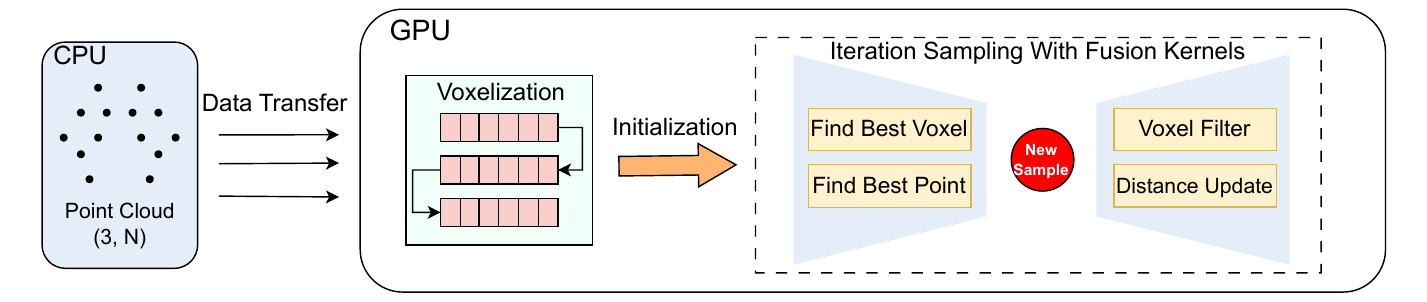}
    \caption{Architectural workflow of our proposed GPU-based RadiusFPS (RadiusFPS-G). Following data transfer to GPU global memory, the input point cloud is voxelized on-device into an optimized, sorted data layout organized by voxel indices. In the iterative sampling process, we strategically fuse four core operations (Find Best Voxel, Find Best Point, Voxel Filter, Distance Update) into two paired fusion kernels to minimize kernel launch overhead and maximize parallel efficiency. The only synchronization point between these kernels is the acquisition of the next sampled point in each iteration.}
    \label{fig:gracc}
\end{figure*}

\section{RadiusFPS-G: GPU-based RadiusFPS Design}
To further improve RadiusFPS on GPUs, we design RadiusFPS-G, a GPU-resident implementation that exploits massive thread-level parallelism and coalesced memory access.
From the RadiusFPS introduction in Sec.\ref{sec:method}, we can observe that although sampled points in the sampling process show dependencies, each component of the algorithm is independent of the others, which results in frequent memory access.

The architectural disparity between the CPU and GPU lies primarily in how they handle memory hierarchy. Unlike the CPU, which relies on large, multi-level caches to reduce latency, the GPU hides latency through hardware multithreading. However, the performance of the GPU is strictly bound by its memory access patterns. In the context of RadiusFPS, frequent global memory transactions and high-frequency communication between independent modules lead to 'memory-wall' issues, necessitating a specialized accelerator design to optimize data movement.

To address these architectural bottlenecks, we propose the RadiusFPS-G. As illustrated in Fig.\ref{fig:gracc}, RadiusFPS-G offloads the entire sampling pipeline to the GPU to minimize host-device synchronization. Following initial voxelization and memory allocation, the core innovation of RadiusFPS-G is the deployment of Fusion Kernels during the iterative sampling phase.
While RadiusFPS is inherently sequential due to its iterative nature, the internal sub-tasks—including Best Voxel Selection, Point Searching, Voxel Filtering, and Distance Updating—are traditionally executed as discrete operations. By fusing these into high-performance kernels, we maintain intermediate data within on-chip shared memory and registers. This strategy eliminates redundant round-trips to the high-latency global memory, transforming the algorithm from a memory-bound process into a compute-efficient pipeline. In this section, we will introduce the Voxelization and Initialization, Fusion Kernel 1: Voxel and Point Selection, Fusion Kernel 2: Voxel Filter and Distance Update.

\begin{algorithm}[p]
\caption{Active Voxelization and Global Initialization on GPU}
\label{alg:voxelization_gpu}
\begin{algorithmic}[1]

\REQUIRE Raw Point Cloud $\mathcal{P}_{AoS} \in \mathbb{R}^{N \times 3}$, Voxel Resolution $n_{\mathrm{vox}}$ (number of voxel bins per axis)
\ENSURE Sorted Point Cloud $\mathcal{P}_{SoA}$, Active Voxel List $\mathcal{V}_{active}$, Compact Voxel Offsets $\mathcal{O}$, Voxel Counts $\mathcal{C}$, Point Distances $\mathcal{D}_p$, Voxel Bounds $\mathcal{D}_v$

\STATE Compute bounding box $[p_{min}, p_{max}]$ of $\mathcal{P}_{AoS}$
\STATE Calculate voxel size $s \leftarrow \frac{\alpha\max(p_{max}-p_{min})}{n_{\mathrm{vox}}}$, where $\alpha=1+\varepsilon$
\STATE Initialize Voxel IDs array $V_{id}$ and Original Indices array $I_{orig}$ of size $N$

\STATE
\STATE \textbf{Phase 1: Parallel Voxel Mapping}

\FORALL{point index $i \in \{0,\dots,N-1\}$ \textbf{in parallel}}
    \STATE Compute voxel coordinates $(v_x,v_y,v_z) \leftarrow \operatorname{clip}(\lfloor \frac{\mathcal{P}_{AoS}[i]-p_{min}}{s} \rfloor,0,n_{\mathrm{vox}}-1)$
    \STATE $V_{id}[i] \leftarrow v_z n_{\mathrm{vox}}^2 + v_y n_{\mathrm{vox}} + v_x$
    \STATE $I_{orig}[i] \leftarrow i$
\ENDFOR

\STATE $(V_{id}, I_{orig}) \leftarrow \text{ParallelRadixSortByKey}(V_{id}, I_{orig})$

\STATE
\STATE \textbf{Phase 2: Gather and AoS-to-SoA Transformation}

\FORALL{point index $i \in \{0,\dots,N-1\}$ \textbf{in parallel}}
    \STATE $idx \leftarrow I_{orig}[i]$ \COMMENT{Fetch original index}
    \STATE $\mathcal{P}_{SoA}[i] \leftarrow \mathcal{P}_{AoS}[idx].x$
    \STATE $\mathcal{P}_{SoA}[i+N] \leftarrow \mathcal{P}_{AoS}[idx].y$
    \STATE $\mathcal{P}_{SoA}[i+2N] \leftarrow \mathcal{P}_{AoS}[idx].z$
\ENDFOR

\STATE
\STATE \textbf{Phase 3: Active Voxel Compaction and Offset Calculation}

\STATE $\mathcal{V}_{active} \leftarrow \text{ParallelUnique}(V_{id})$
\STATE Initialize compact Offsets $\mathcal{O}$ and Counts $\mathcal{C}$ with size $|\mathcal{V}_{active}|$

\FORALL{active-voxel index $k \in \{0,\dots,|\mathcal{V}_{active}|-1\}$ \textbf{in parallel}}
    \STATE $vid \leftarrow \mathcal{V}_{active}[k]$
    \STATE $idx_{start} \leftarrow \text{LowerBound}(V_{id},vid)$
    \STATE $idx_{end} \leftarrow \text{UpperBound}(V_{id},vid)$
    \STATE $\mathcal{O}[k] \leftarrow idx_{start}$
    \STATE $\mathcal{C}[k] \leftarrow idx_{end}-idx_{start}$
\ENDFOR

\STATE
\STATE \textbf{Phase 4: Global Distance Initialization}

\STATE Randomly select initial seed point $p_{seed} \in \mathcal{P}_{SoA}$
\STATE Initialize Point Distances $\mathcal{D}_p$ of size $N$ with $+\infty$
\STATE Initialize Voxel Bounds $\mathcal{D}_v$ of size $|\mathcal{V}_{active}|$ with $0$

\FORALL{point index $i \in \{0,\dots,N-1\}$ \textbf{in parallel}}
    \STATE $\mathcal{D}_p[i] \leftarrow \|\mathcal{P}_{SoA}[i]-p_{seed}\|_2$
\ENDFOR

\FORALL{active-voxel index $k \in \{0,\dots,|\mathcal{V}_{active}|-1\}$ \textbf{in parallel}}
    \STATE $idx_{start} \leftarrow \mathcal{O}[k]$
    \STATE $idx_{end} \leftarrow \mathcal{O}[k]+\mathcal{C}[k]-1$
    \STATE $\mathcal{D}_v[k] \leftarrow \text{BlockReduceMax}(\mathcal{D}_p[idx_{start}\dots idx_{end}])$
\ENDFOR

\RETURN $\mathcal{P}_{SoA},\mathcal{V}_{active},\mathcal{O},\mathcal{C},\mathcal{D}_p,\mathcal{D}_v$

\end{algorithmic}
\end{algorithm}

\subsection{Voxelization and Initialization}
While previous studies on GPU-accelerated point cloud processing provide foundational references for voxel construction \cite{van_cpu,van_gpu}, their reliance on variable locks to manage voxel states and prevent thread contention introduces significant synchronization overhead. This locking strategy cannot be directly applied to RadiusFPS-G. Instead, to maximize parallel efficiency, our architecture relies on the concept of active voxels. By explicitly pruning empty voxels and contiguously packing the populated ones in memory, our structure avoids thread collisions while unlocking the massive bandwidth advantages of memory coalescing. As outlined in Algorithm \ref{alg:voxelization_gpu}, our voxelization strategy is divided into four primary stages: parallel voxel mapping, AoS-to-SoA transformation, sparse structure construction, and global initialization.

\textbf{Parallel Voxel Mapping.}
 To establish spatial locality without relying on thread locks, a parallel GPU kernel independently maps each 3D point to a discrete voxel. The coordinates $(v_x, v_y, v_z)$ are clamped to the index range $[0,n_{\mathrm{vox}}-1]$, where $n_{\mathrm{vox}}$ is the voxel-count resolution, flattened into a unique 1D voxel identifier $V_{id}[i]$, and the original memory index $I_{orig}[i] = i$ is recorded.
 \begin{equation}
     V_{id}[i] = v_z \times n_{\mathrm{vox}}^2 + v_y \times n_{\mathrm{vox}} + v_x
 \end{equation}
However, since spatially adjacent points may still be scattered in global memory, we apply a parallel Radix Sort using $V_{id}$ as keys. With an $O(N)$ linear time complexity, Radix Sort efficiently reorders $I_{orig}$, packing points from the same physical voxel into contiguous memory addresses. This operation fundamentally transforms spatial proximity into physical memory locality, preparing the data for coalesced memory access.

\textbf{Gather and AoS-to-SoA Transformation.}
While sorted indices establish logical locality, fetching data from the original Array of Structures (AoS, e.g., $x_1, y_1, z_1, x_2 \dots$) layout still causes uncoalesced memory access due to coordinate striding. To eliminate this hardware bottleneck, we utilize a dedicated gather kernel to transfer the point cloud into a Structure of Arrays (SoA, e.g., $x_1, x_2 \dots, y_1, y_2 \dots$) layout in parallel. This critical conversion guarantees strictly coalesced memory transactions for GPU thread warps, maximizing global memory bandwidth utilization during neighborhood queries.

\textbf{Sparsity and Offset Calculation.}
To prevent wasting computational resources on empty volumetric space, RadiusFPS-G explicitly extracts an active voxel list. By calculating the boundary offsets—specifically the start index and point count—for each populated voxel, we construct a compact, sparse representation of the voxel. This sparse structure ensures that the subsequent iterative sampling process exclusively evaluates populated regions, avoiding redundant distance calculations and maximizing overall execution throughput.

\textbf{Global Distance Initialization.}
To bootstrap the iterative sampling loop, RadiusFPS-G must perform an initial global distance evaluation. Unlike the CPU implementation, which relies on a straightforward linear scan, the GPU must carefully avoid control-flow divergence to maintain high execution efficiency. After selecting the initial seed point, we launch a dedicated, branchless update kernel. Because the point cloud is now strictly formatted in the Structure of Arrays (SoA) layout, memory transactions for extracting the $X$, $Y$, and $Z$ coordinates are coalesced. For every point $p_i$, the kernel computes the Euclidean distance to the seed point without utilizing divergent branch instructions, thus preventing warp divergence within the Streaming Multiprocessors (SMs). Concurrently, to initialize the voxel-level bounds required for our pruning strategy, RadiusFPS-G performs a parallel block reduction over each compact active-voxel segment. Utilizing fast on-chip shared memory, threads within the same CUDA block collaboratively determine the maximum distance within their assigned voxel segment, establishing the global distance state without mixing raw voxel IDs and compact active-voxel indices.

\begin{figure}[t]
    \centering
    \includegraphics[width=0.6\linewidth]{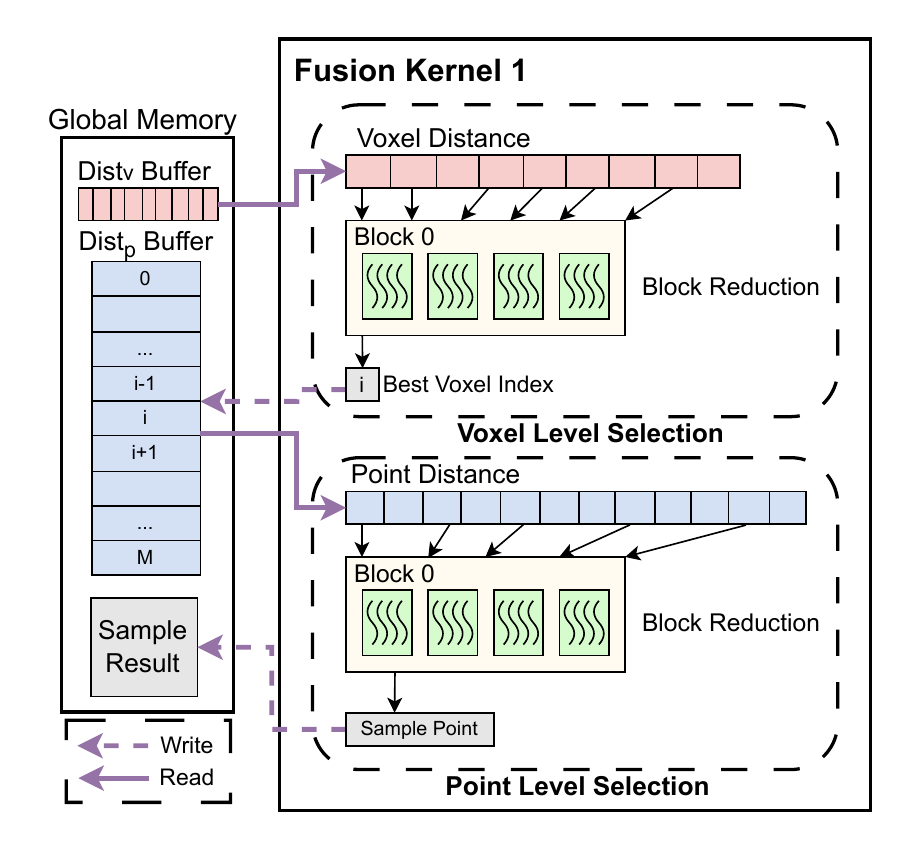}
    \caption{Overview of Fusion Kernel 1. The kernel fuses two sequential parallel reductions into a single operation. First, a block reduction on the $Dist_v$ Buffer determines the best voxel index $i$. This index is subsequently used to query the $Dist_p$ Buffer, where a second block reduction identifies the final Sample Point within that specific voxel. The result is then written directly to the Sample Result buffer in global memory, eliminating unnecessary kernel launch overhead.}
    \label{fig:fusionkernel1}
\end{figure}

\begin{figure}[t]
    \centering
    \includegraphics[width=0.6\linewidth]{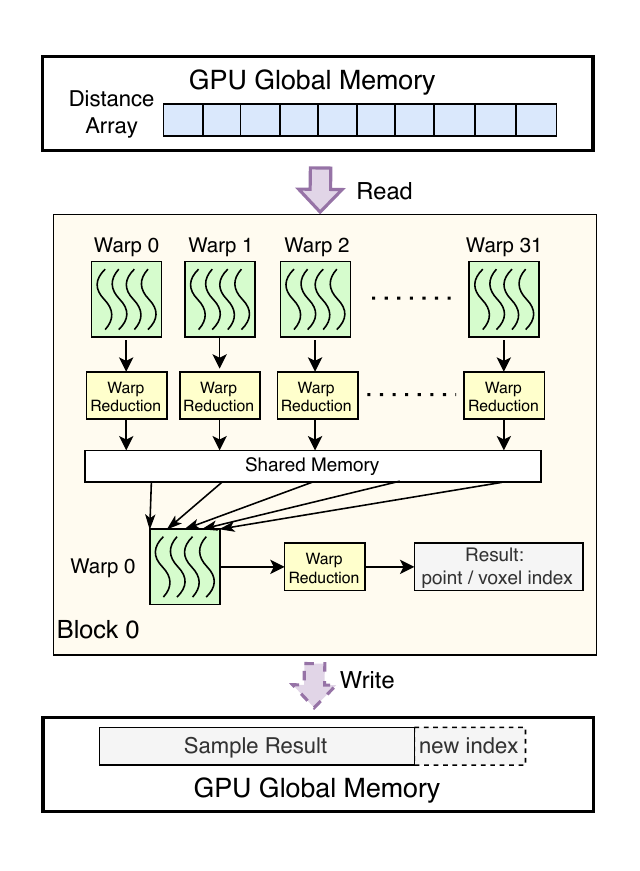}
    \caption{Illustration of the two-stage warp-level reduction within a single CUDA block. To minimize synchronization overhead, each warp first performs an independent intra-warp reduction on data fetched from global memory, storing partial results in shared memory. Subsequently, a single active warp (Warp 0) performs a final reduction on these intermediate values to determine the optimal point or voxel index, writing the result back to global memory.}
    \label{fig:blockreduction}
\end{figure}

\subsection{Fusion Kernel 1: 2-Level Reduction With Efficient Memory Access}

To mitigate the hardware bottlenecks caused by frequent large-scale reductions and excessive global memory accesses, we propose Fusion Kernel 1. As illustrated in the architectural overview (Fig.\ref{fig:fusionkernel1}), this kernel fundamentally optimizes the farthest point selection process by decomposing it into a hierarchical, two-stage pipeline: Voxel Level Selection and Point Level Selection.
By fusing these operations, the kernel completes the farthest point selection with only a few instances of coalesced memory access. The workflow of the kernel operates as follows:

\textbf{Voxel Level Selection}: As shown in the upper section of Fig.~\ref{fig:fusionkernel1}, the kernel initiates the process by reading the voxel distance bounds from the $Dist_v$ Buffer stored in the Global Memory. Within the GPU streaming multiprocessors, a thread block (Block 0) executes a highly parallelized warp-level reduction across these voxel distances.
Fig.~\ref{fig:blockreduction} illustrates the detailed workflow of the warp-level reduction. After fetching the voxel distance bounds ($Dist_v$) from the global memory, the data is distributed among the threads within individual warps. Following an initial intra-warp reduction via register shuffles, the local maximum distance identified by each warp is written into the on-chip shared memory.
\begin{algorithm}[H]
\caption{Warp-Level Argmax Reduction}
\label{alg:warp_argmax}
\begin{algorithmic}[1]

\REQUIRE Each thread holds a pair $(d,-i_{orig})$, where $d$ is a distance value
\ENSURE All threads in the warp obtain the maximum pair under deterministic tie-breaking

\FOR{$k = \log_2(W)-1$ down to $0$}
    \STATE $offset \gets 2^k$
    \STATE $(d', -i'_{orig}) \gets \textsc{ShuffleDown}((d,-i_{orig}), offset)$
    \STATE $(d,-i_{orig}) \gets \arg\max\{(d,-i_{orig}),(d',-i'_{orig})\}$
\ENDFOR

\end{algorithmic}
\end{algorithm}
The specific operation of the warp reduction is introduced in Alg.~\ref{alg:warp_argmax}. The operation is implemented using shuffle-down intrinsics.
The warp reduction is implemented using a recursive halving approach. At each step, threads communicate with a partner at a distance that is halved successively, exploiting shuffle-down intrinsics to aggregate the maximum values.
Subsequently, in the final stage, these intermediate maximums are fed exclusively into Warp 0. Warp 0 then performs a final pass of warp reduction to efficiently determine the absolute maximum for the entire block, thereby completing the reduction process.

\textbf{Point Level Selection}: Transitioning to the lower section of the pipeline, the kernel utilizes the newly identified Best Voxel Index $k$ to precisely target the corresponding compact active-voxel segment. It fetches the exact point distances from the $Dist_p$ Buffer in the Global Memory. Block 0 then performs a secondary block reduction exclusively on the points residing within this optimal voxel, again using the original-index tie-breaking rule described in Sec.~\ref{sec:method}.

\begin{figure}[h]
    \centering
    \includegraphics[width=0.55\linewidth]{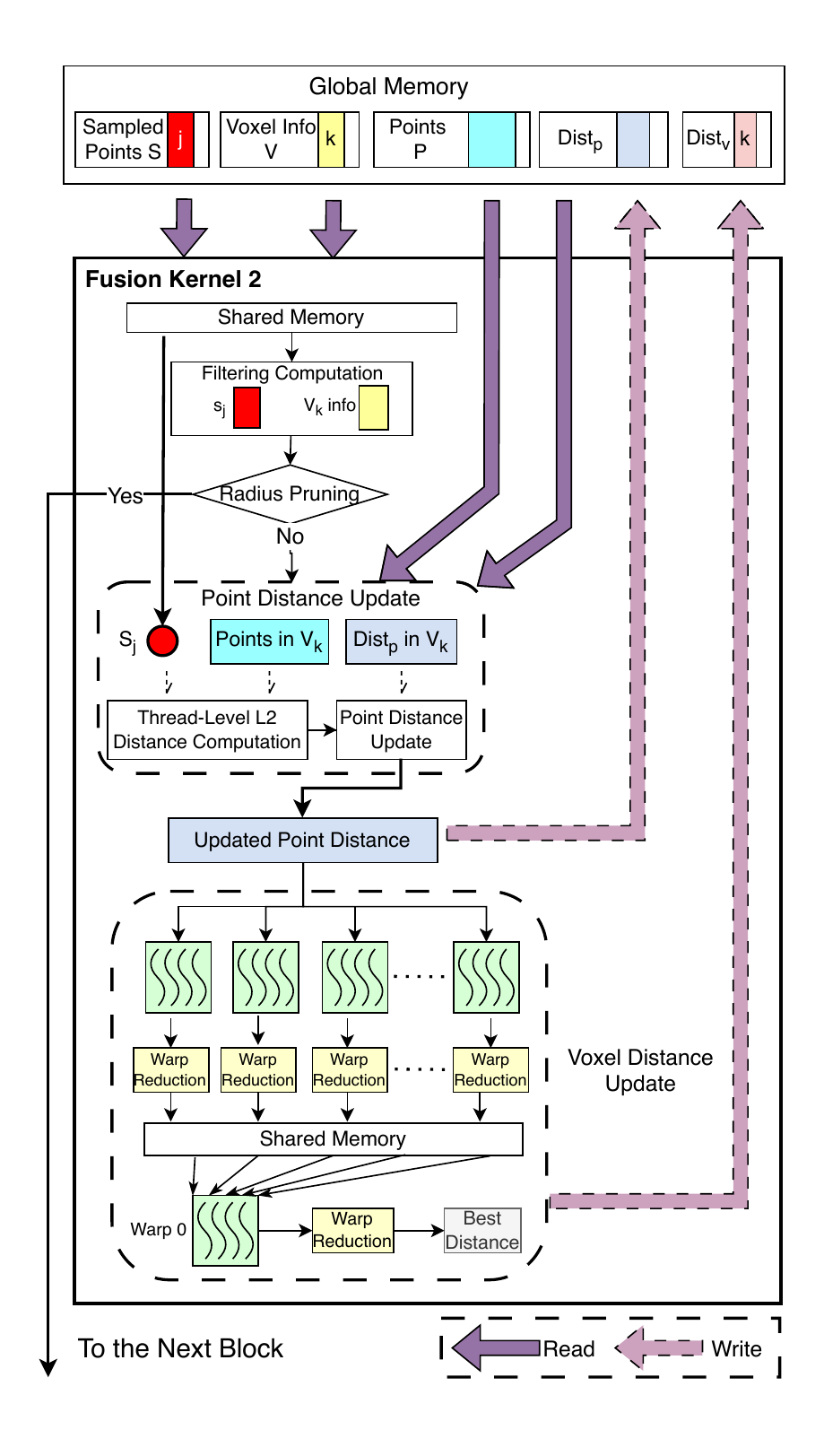}
    \caption{Execution pipeline of Fusion Kernel 2. This kernel efficiently integrates radius-based voxel pruning with hierarchical distance updates. By evaluating the filtering condition early, the kernel dynamically bypasses redundant point-level Euclidean distance computations for pruned voxels. For unpruned voxels, it performs parallel thread-level point distance updates, immediately followed by an optimized warp-level reduction to update the global voxel distance ($Dist_v$), effectively minimizing global memory bandwidth consumption.}
    \label{fig:fusionkernel2}
\end{figure}

\subsection{Fusion Kernel 2: Voxel Filter and Distance Update}
As illustrated in Fig.\ref{fig:fusionkernel2}, Fusion Kernel 2 is responsible for the concurrent execution of voxel-level filtering, point-level distance updating, and voxel state synchronization. A fundamental architectural distinction between Fusion Kernel 1 and Fusion Kernel 2 lies in their workload distribution strategies. While Kernel 1 performs a global two-level reduction to pinpoint a single optimal voxel candidate, Kernel 2 is designed for massive parallel updating across the entire point cloud. Therefore, we adopt a "One-Block-to-One-Voxel" mapping paradigm in the kernel design. The active voxel information is distributed across multiple CUDA blocks, allowing each block to independently process a specific spatial region without cross-block dependency. By consolidating these sequential operations into a single kernel execution, we effectively eliminate the overhead of intermediate global memory round-trips.

Before execution, the block first loads the newly sampled point $s_j$ and the information of its assigned voxel $v_k$ (including its centroid and radius) into low-latency shared memory. Utilizing this shared data, the block collectively performs the filtering computation. Crucially, the GPU execution strictly implements the radius pruning strategy based on spherical voxels introduced in Sec.~\ref{sec:radiuspruning}, and the block evaluates the same pruning condition defined in Eq.~\eqref{eq:radius_pruning_condition}. If $LB(v_k,s_j)$ is no smaller than the currently recorded maximum distance $Dist_v[k]$, the pruning condition is met. Consequently, the entire thread block executes an early exit, freeing up the Streaming Multiprocessor (SM) to schedule the next block immediately without fetching the underlying point data.


\begin{algorithm}[h]
\caption{Voxel Filter and Point Distance Update with Coalesced Access}
\label{alg:pointupdate}
\begin{algorithmic}[1]

\REQUIRE Active-voxel index $k$, newly sampled point $s_j$, point cloud in SoA layout $\mathcal{P}_{SoA}$ with size $N$, compact offsets $\mathcal{O}$, counts $\mathcal{C}$, point distances $\mathcal{D}_p$, voxel bounds $\mathcal{D}_v$
\ENSURE Updated $\mathcal{D}_p$ and $\mathcal{D}_v[k]$

\STATE $idx_{start} \leftarrow \mathcal{O}[k]$, $count \leftarrow \mathcal{C}[k]$
\IF{$LB(v_k,s_j) \geq \mathcal{D}_v[k]$}
    \RETURN $\mathcal{D}_v[k]$ \COMMENT{Safe voxel-level pruning}
\ENDIF

\FOR{each thread $tid$ in the block \textbf{in parallel}}
    \STATE $d_{max}^{local} \leftarrow -1.0$
    \FOR{$q \leftarrow tid$; $q < count$; $q \leftarrow q+\text{block\_size}$}
        \STATE $i \leftarrow idx_{start}+q$
        \STATE $p_i^x \leftarrow \mathcal{P}_{SoA}[i]$
        \STATE $p_i^y \leftarrow \mathcal{P}_{SoA}[i + N]$
        \STATE $p_i^z \leftarrow \mathcal{P}_{SoA}[i + 2N]$
        \STATE $d_{old} \leftarrow \mathcal{D}_p[i]$
        \IF{$|p_i^x-s_j^x| \geq d_{old}$ \textbf{or} $|p_i^y-s_j^y| \geq d_{old}$ \textbf{or} $|p_i^z-s_j^z| \geq d_{old}$}
            \STATE $d_{final} \leftarrow d_{old}$ \COMMENT{Coordinate-wise point skip}
        \ELSE
            \STATE $d_{new} \leftarrow \sqrt{(p_i^x - s_j^x)^2 + (p_i^y - s_j^y)^2 + (p_i^z - s_j^z)^2}$
            \STATE $d_{final} \leftarrow \min(d_{old}, d_{new})$
            \IF{$d_{final} < d_{old}$}
                \STATE $\mathcal{D}_p[i] \leftarrow d_{final}$
            \ENDIF
        \ENDIF
        \STATE $d_{max}^{local} \leftarrow \max(d_{max}^{local}, d_{final})$
    \ENDFOR
\ENDFOR
\STATE $\mathcal{D}_v[k] \leftarrow \text{BlockReduceMax}(d_{max}^{local})$
\RETURN $\mathcal{D}_v[k]$

\end{algorithmic}
\end{algorithm}

If the voxel survives the pruning stage, the block proceeds to the Point Distance Update phase, whose detailed execution flow is formalized in Alg.~\ref{alg:pointupdate}. Individual threads within the block fetch the coordinates of points in the compact segment $[\mathcal{O}[k],\mathcal{O}[k]+\mathcal{C}[k])$. Because the point cloud data was reorganized into a Structure of Arrays (SoA) layout during initialization, the threads within a warp can access the X, Y, and Z coordinates with coalesced memory transactions, avoiding the severe performance penalties of strided memory access. Each thread first applies the coordinate-wise skip test from Eq.~\eqref{eq:point_skip_condition}; only points that pass this test require an exact L2 distance computation against $s_j$. The newly selected point itself is included in the same segment, so its distance is updated to zero when processed. The historical minimum distance $\mathcal{D}_p[i]$ is updated only if the newly computed distance provides a tighter geometric bound, which minimizes global memory writes.

To maintain the exactness of the pruning framework for the next iteration, the maximum distance parameter of $V_k$ must be synchronized. Fusion Kernel 2 employs a similar block reduction mechanism introduced in Fig.\ref{fig:blockreduction} to identify the local maximums, and then performs a terminal reduction pass to determine the absolute maximum distance (Best Distance) for the entire voxel. The final value is written back to the $Dist_v[k]$ buffer, concluding the block execution.

\section{Experiments}
\subsection{Experiment Setup}

To rigorously evaluate the performance of RadiusFPS, we test the algorithm on standard point cloud segmentation tasks that typically rely on Farthest Point Sampling (FPS). Specifically, we apply our method to two indoor datasets, S3DIS\cite{s3dis} and ScanNet\cite{scannet}, as well as an outdoor dataset, SemanticKITTI\cite{semantickitti}. Within these settings, we employ two PointNet++-based architectures—PointMetaBase\cite{pointmetabase} and PointVector\cite{pointvector}—to assess the End-to-End latency and sampling quality of RadiusFPS. Furthermore, we integrate RadiusFPS with a learning-based method, FastPoint\cite{fastpoint}, to explore the feasibility of coupling traditional acceleration algorithms with deep learning-based sampling strategies in practical applications.

We measure the End-to-End latency of RadiusFPS on pretrained models by processing all the scenes from the validation set, and utilize Overall Accuracy (OA), mean Intersection Over Union (mIoU) to evaluate the accuracy and quality of sampled point clouds and sampling effect on the following models. For each model, RadiusFPS and other baseline models are applied to the first-layer FPS, which consumes most of the processing time.

We implement CPU-based RadiusFPS and GPU-based RadiusFPS-G and import them into the OpenPoints\cite{openpoints} point cloud segmentation framework. We also import the GPU implementation of FastPoint\cite{fastpoint} and QuickFPS\cite{quickfps} into OpenPoints as baseline methods. The experiment environment setting is summarized in Tab.\ref{tab:env}. The pretrained PointMetaBase\cite{pointmetabase} and PointVector\cite{pointvector} segmentation models are provided by the model zoo of FastPoint\cite{fastpoint}.

\begin{table}[h]
\centering
\caption{Experimental Environment}
\label{tab:env}
\begin{tabular}{l|l}
\hline
\textbf{Hardware} & \textbf{Specification} \\ \hline
CPU & 12th Gen Intel(R) Core(TM) i7-12700KF \\
GPU & NVIDIA RTX 6000 Ada Generation \\
RAM & 32 GB \\
Operating System & Ubuntu 24.04 \\
CUDA Version & 11.6 \\ \hline

\end{tabular}
\end{table}

\begin{table*}[t]
\caption{End-to-End Latency Reduction Evaluation on PointMetaBase}
\label{tab:pointmetabase}
\resizebox{\textwidth}{!}{%
\begin{tabular}{l|ccc|ccc|ccc}
\hline
\textbf{DataSets} &
  \multicolumn{3}{c|}{\textbf{S3DIS}} &
  \multicolumn{3}{c|}{\textbf{ScanNet}} &
  \multicolumn{3}{c}{\textbf{SemanticKITTI}} \\ \hline
\textbf{Methods} &
  \multicolumn{1}{l|}{\textbf{OA (\%)}} &
  \multicolumn{1}{l|}{\textbf{mIoU (\%)}} &
  \multicolumn{1}{l|}{\textbf{Runtime (s)}} &
  \multicolumn{1}{l|}{\textbf{OA (\%)}} &
  \multicolumn{1}{l|}{\textbf{mIoU (\%)}} &
  \multicolumn{1}{l|}{\textbf{Runtime (s)}} &
  \multicolumn{1}{l|}{\textbf{OA (\%)}} &
  \multicolumn{1}{l|}{\textbf{mIoU (\%)}} &
  \multicolumn{1}{l}{\textbf{Runtime (s)}} \\ \hline
FPS (CPU) &
  \multicolumn{1}{c|}{89.59} &
  \multicolumn{1}{c|}{67.90} &
  351.846 &
  \multicolumn{1}{c|}{89.68} &
  \multicolumn{1}{c|}{70.93} &
  2861.794 &
  \multicolumn{1}{c|}{89.53} &
  \multicolumn{1}{c|}{49.72} &
  14697.007 \\
FPS + NPDU &
  \multicolumn{1}{c|}{86.15} &
  \multicolumn{1}{c|}{61.43} &
  89.125 &
  \multicolumn{1}{c|}{82.98} &
  \multicolumn{1}{c|}{55.72} &
  719.155 &
  \multicolumn{1}{c|}{80.67} &
  \multicolumn{1}{c|}{33.97} &
  3947.027 \\
FPS (GPU) &
  \multicolumn{1}{c|}{89.60} &
  \multicolumn{1}{c|}{67.90} &
  40.176 &
  \multicolumn{1}{c|}{89.63} &
  \multicolumn{1}{c|}{70.80} &
  296.860 &
  \multicolumn{1}{c|}{89.59} &
  \multicolumn{1}{c|}{49.71} &
  1734.057 \\
QuickFPS &
  \multicolumn{1}{c|}{89.60} &
  \multicolumn{1}{c|}{67.90} &
  18.363 &
  \multicolumn{1}{c|}{89.60} &
  \multicolumn{1}{c|}{70.96} &
  106.466 &
  \multicolumn{1}{c|}{89.54} &
  \multicolumn{1}{c|}{49.54} &
  859.166 \\
\textbf{RadiusFPS} &
  \multicolumn{1}{c|}{89.60} &
  \multicolumn{1}{c|}{67.90} &
  15.948 &
  \multicolumn{1}{c|}{89.63} &
  \multicolumn{1}{c|}{70.80} &
  86.307 &
  \multicolumn{1}{c|}{\textbf{89.59}} &
  \multicolumn{1}{c|}{\textbf{49.71}} &
  1065.902 \\
\textbf{RadiusFPS-G} &
  \multicolumn{1}{c|}{\textbf{89.75}} &
  \multicolumn{1}{c|}{\textbf{68.28}} &
  15.335 &
  \multicolumn{1}{c|}{\textbf{89.75}} &
  \multicolumn{1}{c|}{\textbf{71.21}} &
  92.346 &
  \multicolumn{1}{c|}{89.58} &
  \multicolumn{1}{c|}{49.26} &
  691.230 \\ \hline
FastPoint + FPS &
  \multicolumn{1}{c|}{89.53} &
  \multicolumn{1}{c|}{68.21} &
  14.297 &
  \multicolumn{1}{c|}{89.69} &
  \multicolumn{1}{c|}{70.87} &
  89.009 &
  \multicolumn{1}{c|}{89.44} &
  \multicolumn{1}{c|}{49.26} &
  622.247 \\
FastPoint + QuickFPS &
  \multicolumn{1}{c|}{89.60} &
  \multicolumn{1}{c|}{68.28} &
  13.083 &
  \multicolumn{1}{c|}{89.67} &
  \multicolumn{1}{c|}{70.96} &
  69.475 &
  \multicolumn{1}{c|}{89.44} &
  \multicolumn{1}{c|}{49.26} &
  572.870 \\
\textbf{FastPoint + RadiusFPS} &
  \multicolumn{1}{c|}{89.56} &
  \multicolumn{1}{c|}{68.19} &
  12.308 &
  \multicolumn{1}{c|}{89.11} &
  \multicolumn{1}{c|}{69.38} &
  69.947 &
  \multicolumn{1}{c|}{88.75} &
  \multicolumn{1}{c|}{47.48} &
  563.614 \\
\textbf{FastPoint + RadiusFPS-G} &
  \multicolumn{1}{c|}{89.25} &
  \multicolumn{1}{c|}{67.58} &
  \textbf{11.248} &
  \multicolumn{1}{c|}{89.08} &
  \multicolumn{1}{c|}{69.43} &
  \textbf{68.860} &
  \multicolumn{1}{c|}{88.75} &
  \multicolumn{1}{c|}{47.55} &
  \textbf{531.202} \\ \hline
\end{tabular}%
}
\end{table*}

\begin{table*}[t]
\caption{End-to-End Latency Reduction Evaluation on PointVector}
\label{tab:pointvector}
\resizebox{\textwidth}{!}{%
\begin{tabular}{l|ccc|ccc|ccc}
\hline
\textbf{DataSets} &
  \multicolumn{3}{c|}{\textbf{S3DIS}} &
  \multicolumn{3}{c|}{\textbf{ScanNet}} &
  \multicolumn{3}{c}{\textbf{SemanticKITTI}} \\ \hline
\textbf{Method} &
  \multicolumn{1}{l|}{\textbf{OA (\%)}} &
  \multicolumn{1}{l|}{\textbf{mIoU (\%)}} &
  \multicolumn{1}{l|}{\textbf{Runtime (s)}} &
  \multicolumn{1}{l|}{\textbf{OA (\%)}} &
  \multicolumn{1}{l|}{\textbf{mIoU (\%)}} &
  \multicolumn{1}{l|}{\textbf{Runtime (s)}} &
  \multicolumn{1}{l|}{\textbf{OA (\%)}} &
  \multicolumn{1}{l|}{\textbf{mIoU (\%)}} &
  \multicolumn{1}{l}{\textbf{Runtime (s)}} \\ \hline
FPS (CPU) &
  \multicolumn{1}{c|}{90.08} &
  \multicolumn{1}{c|}{69.80} &
  714.228 &
  \multicolumn{1}{c|}{89.54} &
  \multicolumn{1}{c|}{70.21} &
  2780.102 &
  \multicolumn{1}{c|}{89.08} &
  \multicolumn{1}{c|}{48.38} &
  17348.921 \\
FPS + NPDU &
  \multicolumn{1}{c|}{72.23} &
  \multicolumn{1}{c|}{36.32} &
  191.907 &
  \multicolumn{1}{c|}{73.98} &
  \multicolumn{1}{c|}{35.11} &
  739.691 &
  \multicolumn{1}{c|}{74.18} &
  \multicolumn{1}{c|}{26.64} &
  4130.392 \\
FPS (GPU) &
  \multicolumn{1}{c|}{89.96} &
  \multicolumn{1}{c|}{70.03} &
  41.792 &
  \multicolumn{1}{c|}{89.54} &
  \multicolumn{1}{c|}{70.11} &
  319.552 &
  \multicolumn{1}{c|}{89.08} &
  \multicolumn{1}{c|}{48.38} &
  1955.910 \\
QuickFPS &
  \multicolumn{1}{c|}{89.89} &
  \multicolumn{1}{c|}{69.55} &
  22.465 &
  \multicolumn{1}{c|}{89.41} &
  \multicolumn{1}{c|}{69.92} &
  139.196 &
  \multicolumn{1}{c|}{89.07} &
  \multicolumn{1}{c|}{48.25} &
  1074.936 \\
\textbf{RadiusFPS} &
  \multicolumn{1}{c|}{89.96} &
  \multicolumn{1}{c|}{70.03} &
  19.711 &
  \multicolumn{1}{c|}{89.54} &
  \multicolumn{1}{c|}{70.11} &
  114.465 &
  \multicolumn{1}{c|}{89.08} &
  \multicolumn{1}{c|}{48.38} &
  1528.442 \\
\textbf{RadiusFPS-G} &
  \multicolumn{1}{c|}{89.77} &
  \multicolumn{1}{c|}{69.22} &
  19.243 &
  \multicolumn{1}{c|}{89.52} &
  \multicolumn{1}{c|}{69.92} &
  123.651 &
  \multicolumn{1}{c|}{\textbf{89.11}} &
  \multicolumn{1}{c|}{48.28} &
  901.779 \\ \hline
FastPoint + FPS &
  \multicolumn{1}{c|}{89.82} &
  \multicolumn{1}{c|}{69.54} &
  19.207 &
  \multicolumn{1}{c|}{89.54} &
  \multicolumn{1}{c|}{70.04} &
  121.400 &
  \multicolumn{1}{c|}{88.63} &
  \multicolumn{1}{c|}{47.78} &
  791.042 \\
FastPoint + QuickFPS &
  \multicolumn{1}{c|}{89.92} &
  \multicolumn{1}{c|}{69.76} &
  18.337 &
  \multicolumn{1}{c|}{89.50} &
  \multicolumn{1}{c|}{70.04} &
  113.505 &
  \multicolumn{1}{c|}{88.60} &
  \multicolumn{1}{c|}{47.54} &
  738.996 \\
\textbf{FastPoint + RadiusFPS} &
  \multicolumn{1}{c|}{89.40} &
  \multicolumn{1}{c|}{68.59} &
  17.589 &
  \multicolumn{1}{c|}{89.18} &
  \multicolumn{1}{c|}{69.47} &
  109.487 &
  \multicolumn{1}{c|}{87.13} &
  \multicolumn{1}{c|}{45.27} &
  755.050 \\
\textbf{FastPoint + RadiusFPS-G} &
  \multicolumn{1}{c|}{89.46} &
  \multicolumn{1}{c|}{68.31} &
  \textbf{17.278} &
  \multicolumn{1}{c|}{89.18} &
  \multicolumn{1}{c|}{69.34} &
  \textbf{104.636} &
  \multicolumn{1}{c|}{87.15} &
  \multicolumn{1}{c|}{45.28} &
  \textbf{699.001} \\ \hline
\end{tabular}%
}
\end{table*}

\subsection{End-to-End Overall Evaluation on FPS Optimized Strategies}

To comprehensively evaluate the accuracy and efficiency of our proposed RadiusFPS and its GPU-accelerated variant, RadiusFPS-G, we conduct extensive experiments on point cloud segmentation tasks. By integrating these sampling algorithms into the PointMetaBase and PointVector models, we quantify their performance using Overall Accuracy (OA), mean Intersection over Union (mIoU), and total runtime to demonstrate latency reductions. As summarized in Tab. \ref{tab:pointmetabase} and Tab. \ref{tab:pointvector}, we benchmark our methods against standard CPU baselines (FPS and FPS + NPDU) as well as state-of-the-art GPU implementations (GPU-based FPS and QuickFPS). Furthermore, we investigate the synergy between these heuristic methods and FastPoint, a deep learning-based sampling strategy. In these hybrid pipelines, our heuristic methods provide initial spatial priors, which FastPoint subsequently leverages to optimize the predictive sampling process.

The most prominent advantage of our proposed methods lies in the substantial reduction of inference latency, which is essential for large-scale point cloud processing. As shown in Tab. \ref{tab:pointmetabase} and Tab. \ref{tab:pointvector}, the traditional CPU-based FPS suffers from prohibitive computational costs, requiring tens of thousands of seconds on large-scale datasets like SemanticKITTI. While the GPU-accelerated standard FPS (FPS (GPU)) drastically mitigates this issue, our proposed RadiusFPS-G further pushes the boundary of efficiency.
For instance, under the PointMetaBase architecture on the SemanticKITTI dataset, RadiusFPS-G reduces the runtime from 1734.057s (standard GPU FPS) to 691.230s, achieving a speedup of approximately 2.5$\times$ over the GPU baseline. This acceleration trend is consistent across all datasets and both network architectures. Furthermore, when integrating our heuristic samplers with the learning-based FastPoint strategy, the latency reaches its minimum in our experiments. The combination of FastPoint + RadiusFPS-G consistently yields the fastest inference speeds across all evaluated configurations. Notably, on the SemanticKITTI dataset using the PointMetaBase model, it reduces the full validation runtime to 531.202s, indicating substantially improved suitability for latency-sensitive point cloud processing.

Crucially, the computational efficiency of RadiusFPS and its variants is achieved while maintaining competitive segmentation accuracy. As observed in Tab. \ref{tab:pointmetabase}, RadiusFPS-G not only accelerates the process but also achieves the highest Overall Accuracy (OA) and mean Intersection over Union (mIoU) among all standalone samplers on the S3DIS (89.75\% OA, 68.28\% mIoU) and ScanNet (89.75\% OA, 71.21\% mIoU) datasets, while other settings show small metric variations around the GPU-FPS baseline.

While learning-based predictive sampling (FastPoint) trades some accuracy for speed, it remains highly competitive. The FastPoint + RadiusFPS-G pipeline maintains robust metrics (e.g., 88.75\% OA and 47.55\% mIoU on SemanticKITTI in PointMetaBase), with an absolute mIoU drop of about 2.16 percentage points relative to the GPU FPS baseline while reducing runtime by more than two-thirds. In stark contrast, other efficiency-oriented combinations, such as FPS + NPDU, exhibit catastrophic performance degradation, particularly in the PointVector architecture (Tab. \ref{tab:pointvector}), where its mIoU plummets to 35.11\% on ScanNet and 26.64\% on SemanticKITTI. This sharp contrast underscores the robustness of RadiusFPS-G in preserving critical geometric features during downsampling, ensuring that downstream networks receive high-quality representations even under aggressive acceleration.

\begin{figure*}[t]
\centering

\subfloat[]{
\includegraphics[width=0.48\textwidth]{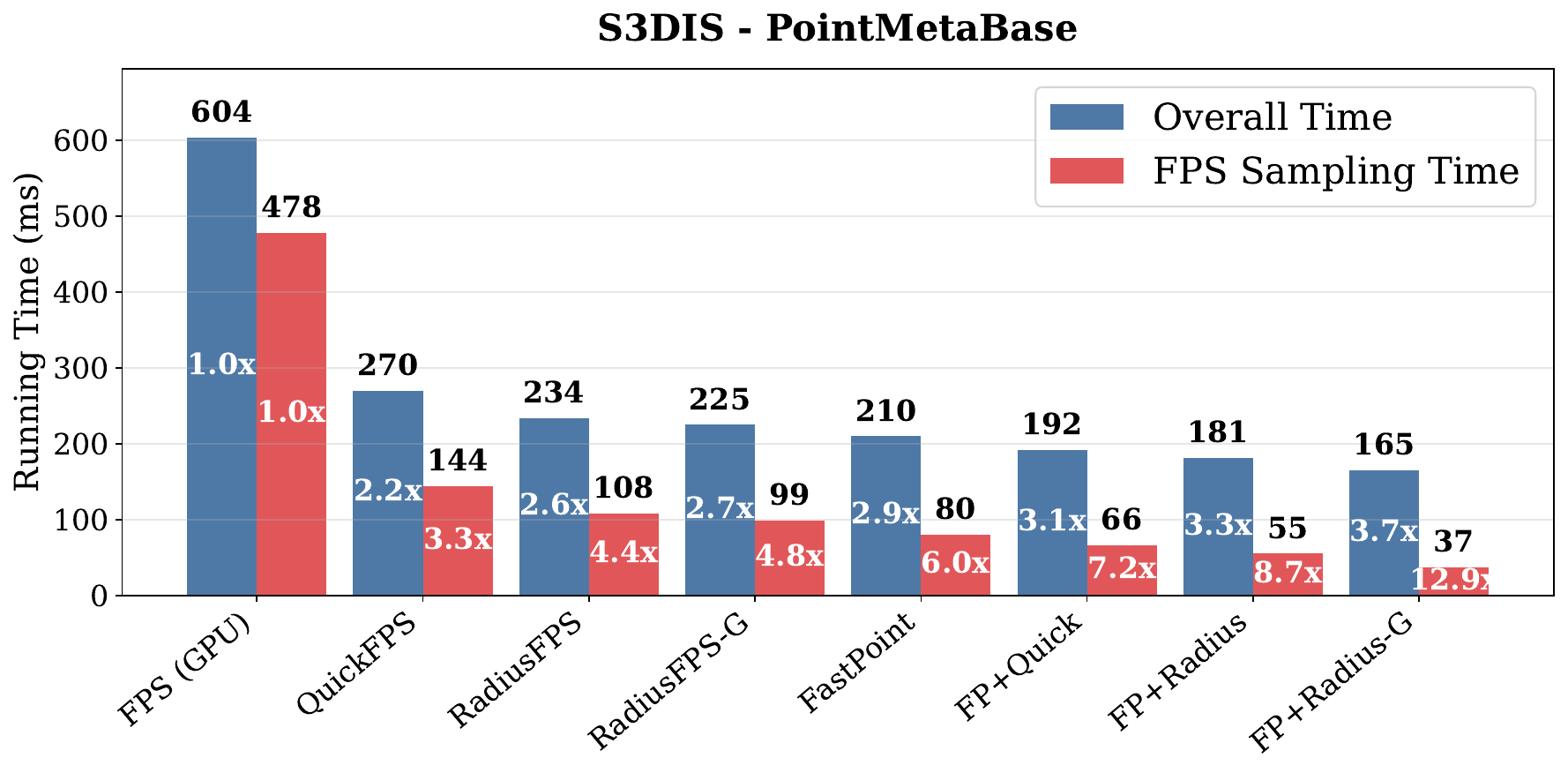}
\label{fig:speedupa}
}
\hfill
\subfloat[]{
\includegraphics[width=0.48\textwidth]{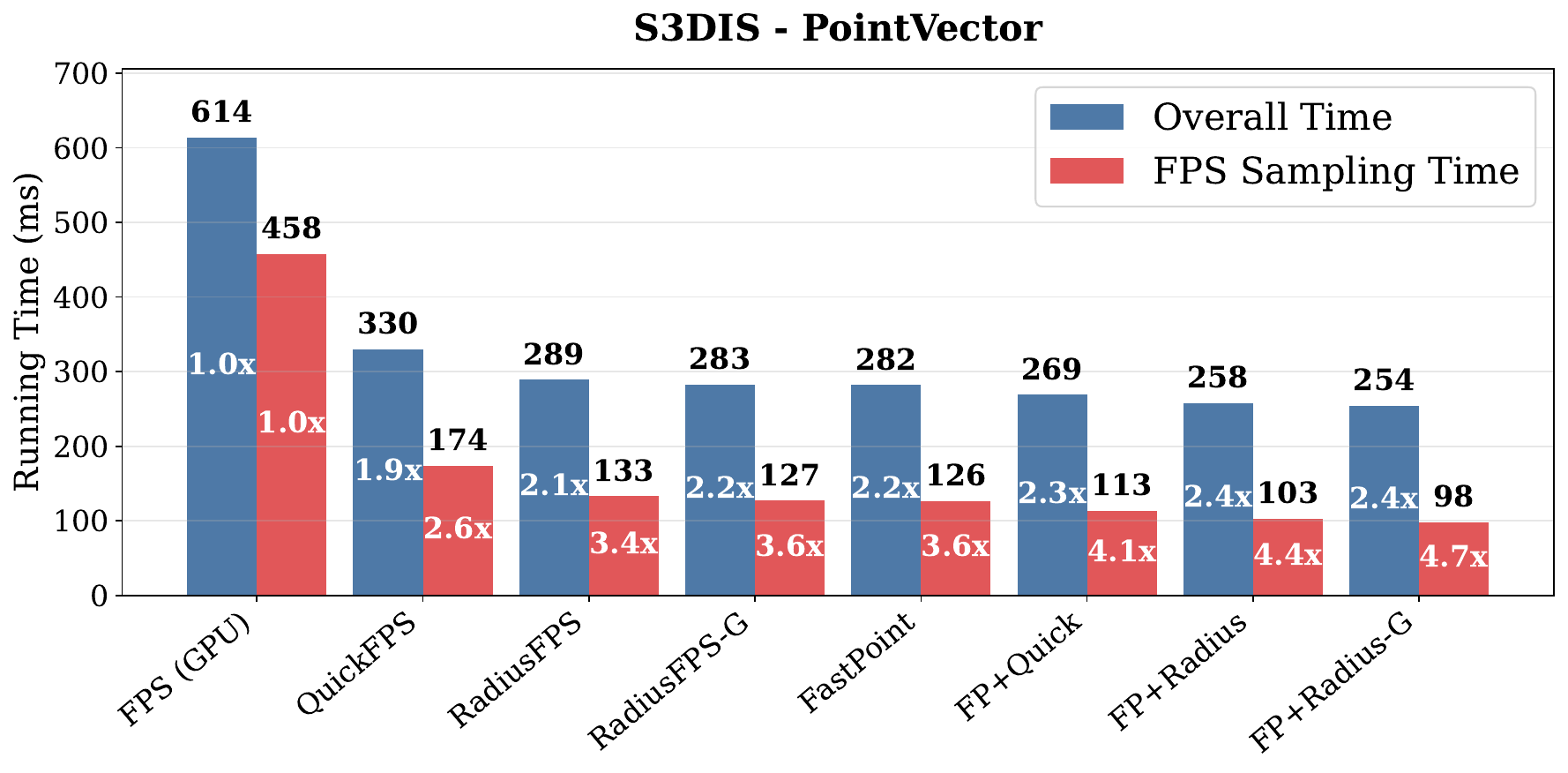}
\label{fig:speedupb}
}

\vspace{3pt} 

\subfloat[]{
\includegraphics[width=0.48\textwidth]{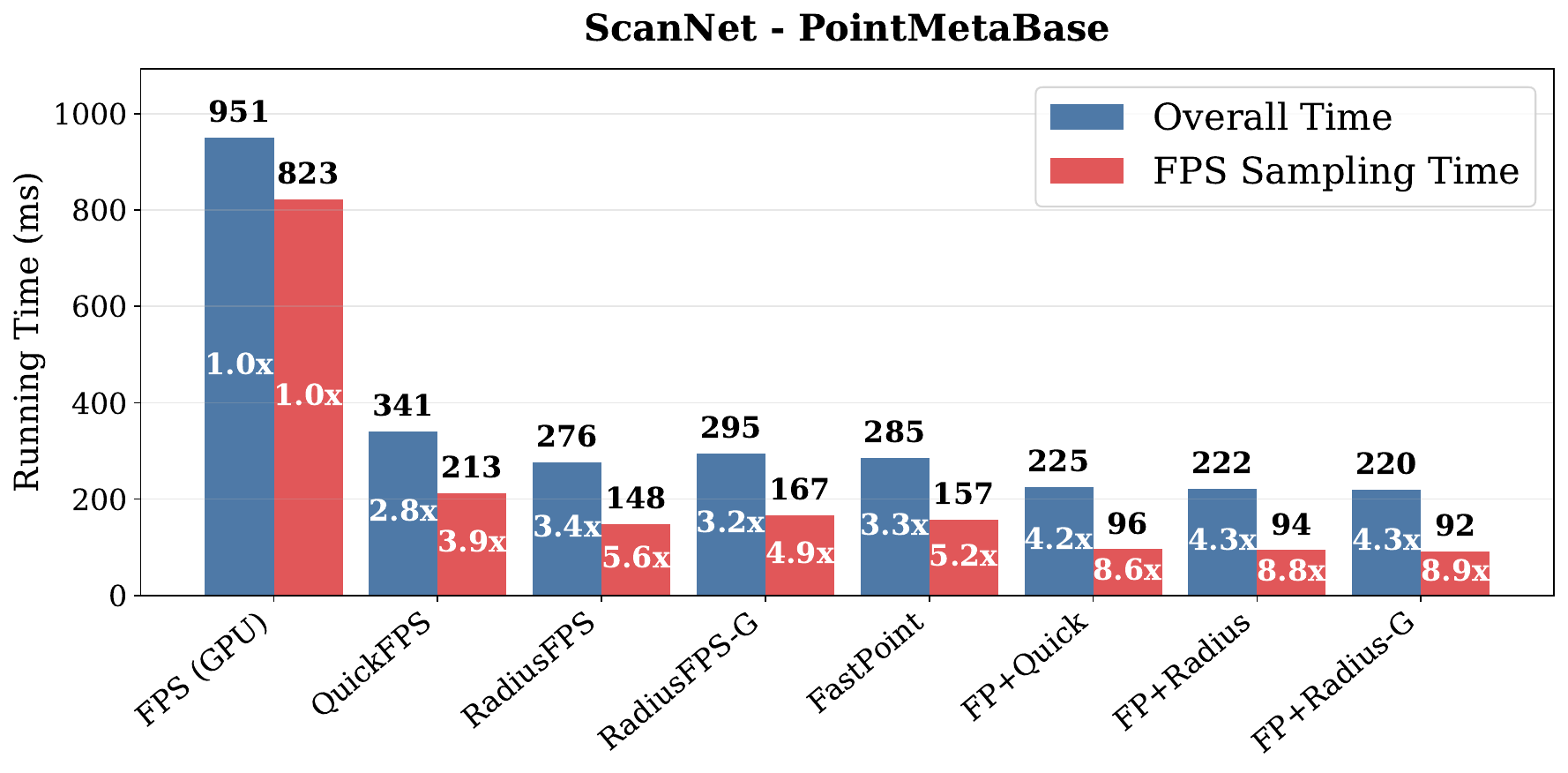}
\label{fig:speedupc}
}
\hfill
\subfloat[]{
\includegraphics[width=0.48\textwidth]{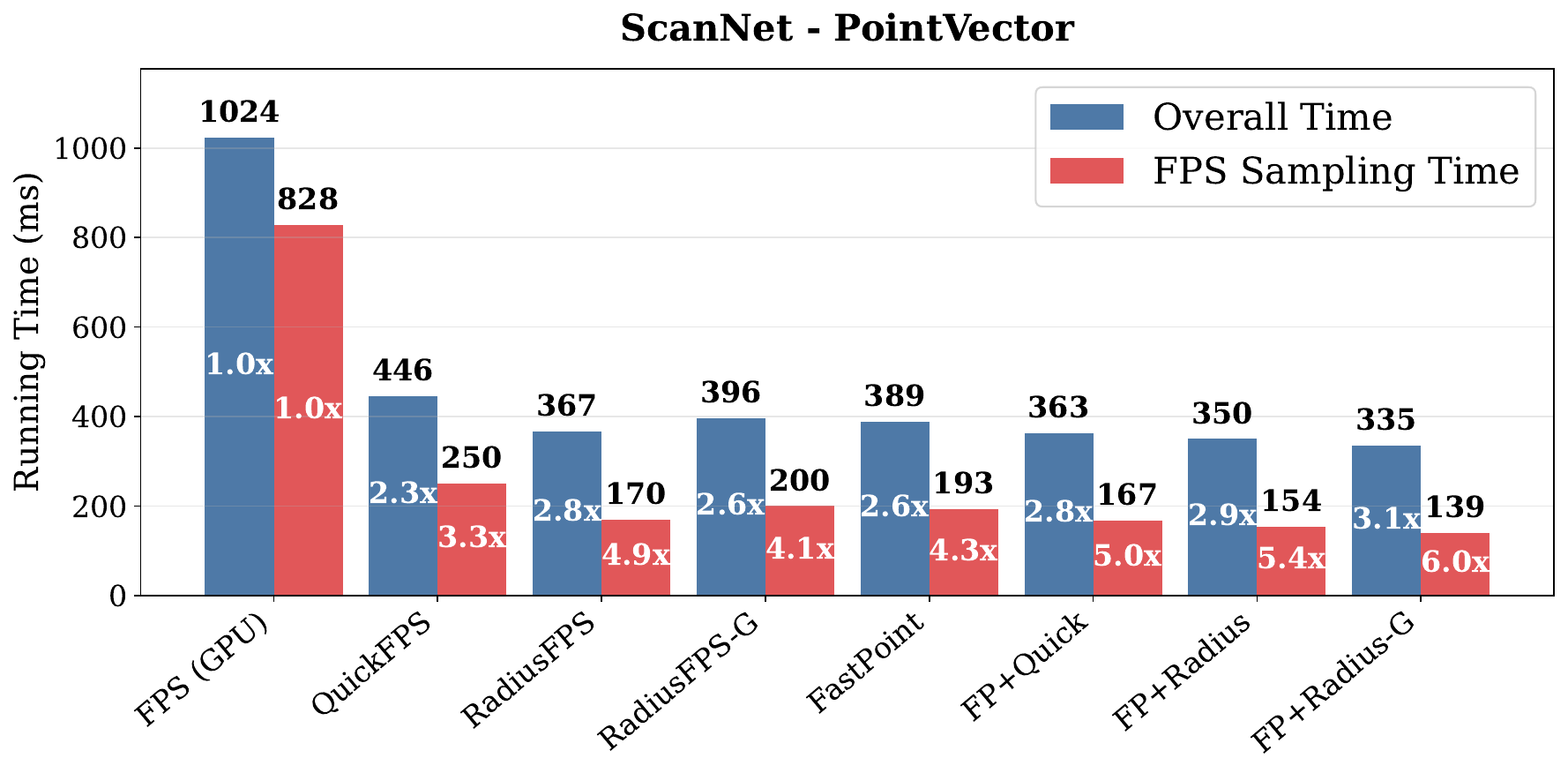}
\label{fig:speedupd}
}

\vspace{3pt} 

\subfloat[]{
\includegraphics[width=0.48\textwidth]{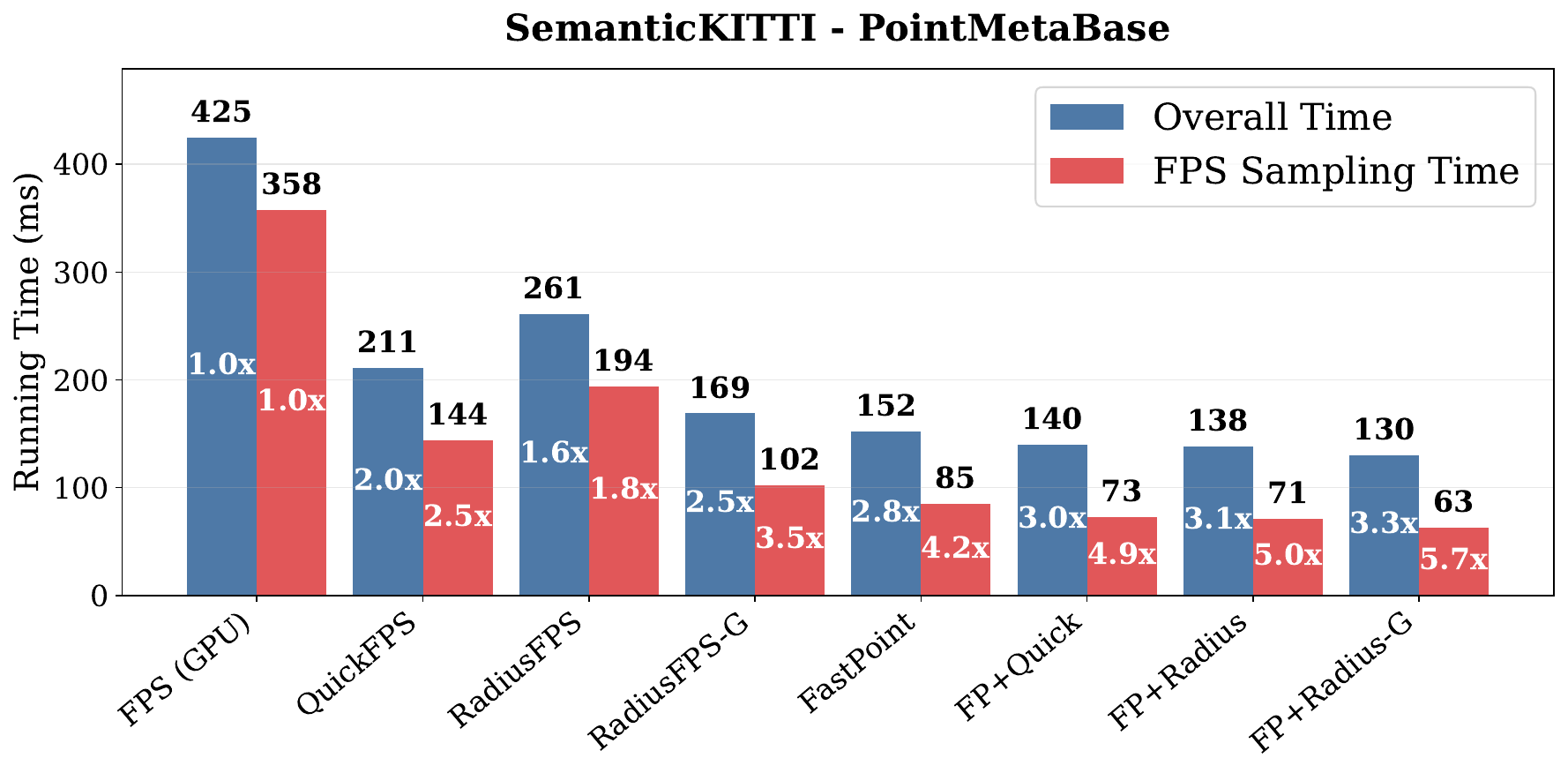}
\label{fig:speedupe}
}
\hfill
\subfloat[]{
\includegraphics[width=0.48\textwidth]{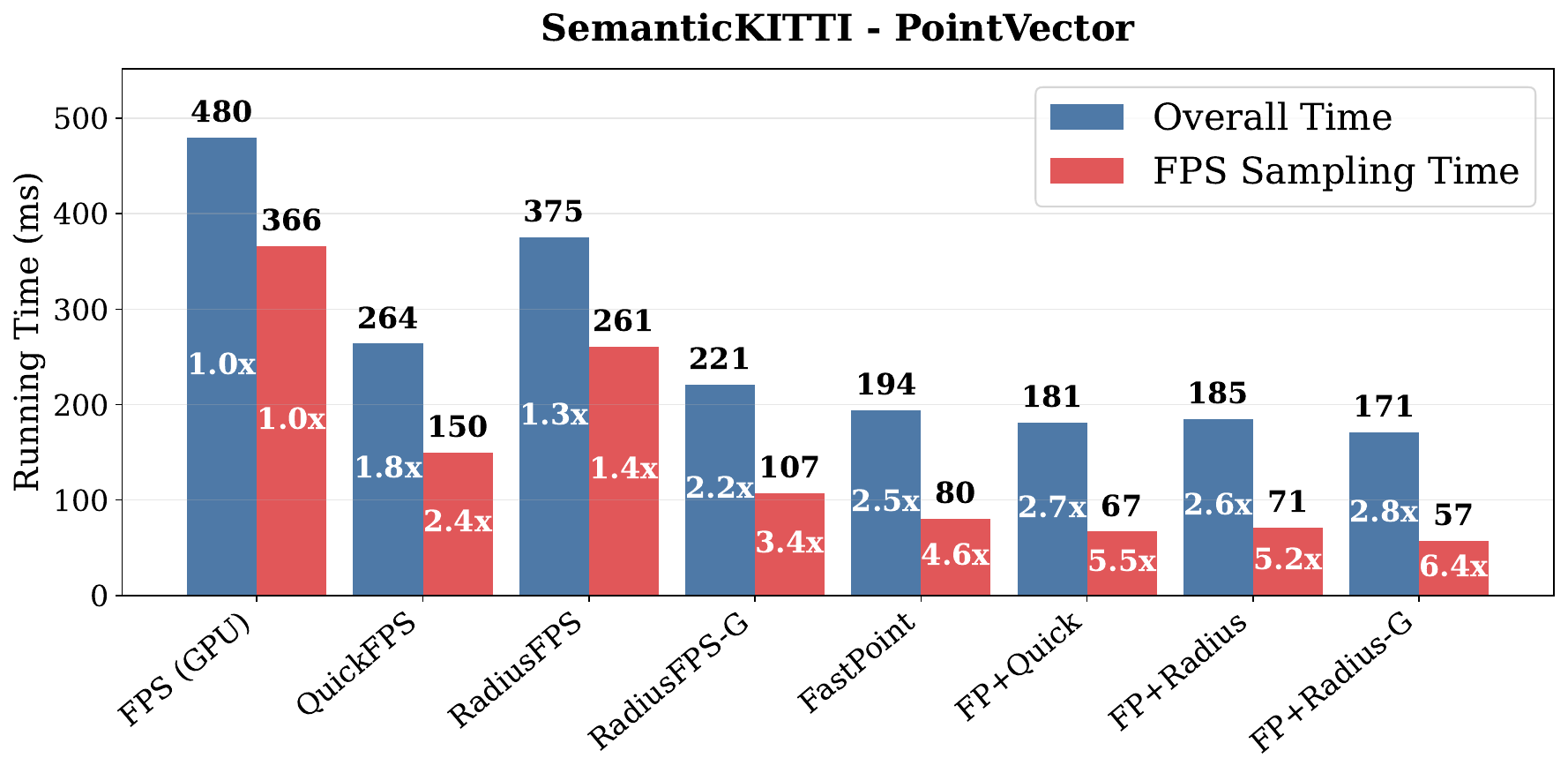}
\label{fig:speedupf}
}

\caption{Comprehensive runtime and speedup analysis across S3DIS, ScanNet, and SemanticKITTI datasets. The figures compare overall inference time (blue) and dedicated sampling time (red) for PointMetaBase (left column) and PointVector (right column) architectures. Annotated values denote the relative speedup compared to the baseline FPS (GPU). Results demonstrate that our proposed RadiusFPS-G significantly reduces latency, while the hybrid pipeline integrating FastPoint with our method (FP+Radius-G) consistently achieves the highest sampling speedup across all evaluated scenarios.}
\label{fig:overall_speedup}
\end{figure*}

\subsection{FPS Efficiency on Different Datasets}
To further dissect the computational efficiency, Fig.~\ref{fig:overall_speedup} visualizes the speedup ratios of our proposed acceleration strategies against the standard GPU-based FPS baseline. We evaluate the performance gain from two perspectives: the End-to-End latency reduction (Overall Time) and the sampling acceleration (FPS Sampling Time).

As illustrated by the darker bars, our proposed sampling variants achieve large speedups over the GPU baseline. Notably, the integration of deep learning-based spatial priors with our heuristic samplers yields the highest efficiency. The FastPoint + RadiusFPS-G combination accelerates the sampling process by up to 11.72$\times$ on the S3DIS dataset with PointMetaBase (Fig.~\ref{fig:speedupa}), demonstrating the potential of this hybrid approach in breaking the traditional FPS bottleneck. While the End-to-End pipeline speedup (lighter bars) is naturally bounded by the constant inference time of other network components (such as convolutions and MLPs), the overall latency reduction remains significant. Across different backbones and datasets, our fastest pipelines deliver a total runtime speedup ranging from approximately 2.4$\times$ to 4.3$\times$ relative to the GPU-FPS baseline. This confirms that the efficiency gains achieved in the sampling stage translate effectively into substantial acceleration for the entire point cloud segmentation task, making it suitable for latency-sensitive applications.

\begin{table}[t]
\centering
\caption{Ablation Study of RadiusFPS and RadiusFPS-G on the S3DIS Dataset}
\label{tab:ablation}
\resizebox{0.6\columnwidth}{!}{%
\begin{tabular}{@{}l|cc|c|c@{}}
\toprule
\textbf{Method} & \multicolumn{2}{c}{\textbf{Configurations}} & \textbf{mIoU (\%)} & \textbf{Total SpeedUp} \\ \midrule
FPS (GPU)  & \multicolumn{2}{c}{-}                       & 67.90              & 1.00$\times$           \\ \midrule
                & \textit{Radius Pruning} & \textit{Point Skip}&                    &                        \\ \cmidrule{2-3}
\multirow{3}{*}{RadiusFPS}
                & \checkmark              & \checkmark         & 67.90              & 2.58$\times$           \\
                & \checkmark              &                    & 67.90              & 2.42$\times$           \\
                &                         & \checkmark         & 19.73              & 3.68$\times$           \\ \midrule
                & \textit{Fusion Kernel 1}& \textit{Fusion Kernel 2}&                 &                        \\ \cmidrule{2-3}
\multirow{4}{*}{RadiusFPS-G}
                & \checkmark              & \checkmark         & 68.28              & 2.68$\times$           \\
                & \checkmark              &                    & 68.28              & 2.24$\times$           \\
                &                         & \checkmark         & 68.28              & 2.11$\times$           \\
                &                         &                    & 68.28              & 1.87$\times$           \\ \bottomrule
\end{tabular}%
}
\end{table}

\subsection{Ablation Studies}
To validate the individual contributions of our proposed algorithmic designs, we first ablate the mechanisms within RadiusFPS. As shown in Tab.\ref{tab:ablation}, applying Radius Pruning alone achieves a significant speedup of 2.42$\times$ without any loss in segmentation accuracy, maintaining the baseline mIoU of 67.90\%. This supports the safety of the conservative voxel-level pruning rule.

Conversely, the standalone Point Skip configuration yields an even higher speedup (3.68$\times$), but results in a severe accuracy drop to 19.73\%. This row should be interpreted as an unguided skip-only negative control rather than the full conservative RadiusFPS update: without voxel-level pruning and synchronization, local coordinate filtering can discard information needed by subsequent selections. When Point Skip is combined with Radius Pruning (the full RadiusFPS), the algorithm achieves a robust 2.58$\times$ speedup while restoring the mIoU to 67.90\%. This demonstrates that Point Skip is effective when it operates under the structural constraints and distance-state synchronization provided by Radius Pruning.
The second half of Tab.\ref{tab:ablation} ablates the hardware-level optimizations implemented in the GPU-accelerated version, RadiusFPS-G. Because these are pure computational optimizations, the segmentation accuracy remains completely unaffected, locked at a stable 68.28\% across all configurations.
The unoptimized GPU baseline (without any fusion kernels) provides a modest 1.87$\times$ speedup over the GPU standard. By introducing Fusion Kernel 1 and Fusion Kernel 2 independently, the total speedup increases to 2.24$\times$ and 2.11$\times$, respectively. This indicates that both kernel optimizations effectively reduce memory access overhead and kernel launch latency. Most importantly, when both fusion strategies are activated simultaneously, the performance synergistically peaks at a 2.68$\times$ total speedup. This confirms that our custom CUDA kernel fusions are orthogonal and complementary, maximizing GPU utilization to break the bottleneck of traditional parallel sampling.

\begin{table*}[t]
\caption{Sampling Quality Evaluation of Segmentation Quality across Various Sampling Strategies}
\label{tab:miou}
\resizebox{\textwidth}{!}{%
\begin{tabular}{|c|c|c|c|c|c|c|c|c|c|}
\hline
Model                          & Datasets & FPS (GPU) & Random & Grid  & EdgePC & FPS + NPDU & FastPoint & RadiusFPS & RadiusFPS-G \\ \hline
\multirow{3}{*}{PointMetaBase} & S3DIS    & 67.90          & 65.28  & 66.81 & 67.16  & 61.43      & 68.21     & 67.90     & 68.28       \\ \cline{2-2}
 & ScanNet       & 70.80 & 64.45 & 69.95 & 70.31 & 50.22 & 70.87 & 70.80 & 71.21 \\ \cline{2-2}
 & SemanticKITTI & 49.72 & 40.90 & 49.26 & 47.45 & 33.97 & 49.26 & 49.71 & 49.26 \\ \hline
\multirow{3}{*}{PointVector}   & S3DIS    & 70.03          & 59.76  & 67.89 & 68.21  & 36.32      & 69.54     & 70.03     & 69.54       \\ \cline{2-2}
 & ScanNet       & 70.21 & 54.65 & 68.90 & 69.02 & 35.11 & 70.04 & 70.11 & 70.04 \\ \cline{2-2}
 & SemanticKITTI & 48.38 & 33.35 & 47.30 & 45.42 & 26.64 & 47.78 & 48.38 & 48.28 \\ \hline
\end{tabular}%
}
\end{table*}

\begin{figure*}[b]
    \centering
    \includegraphics[width=\textwidth]{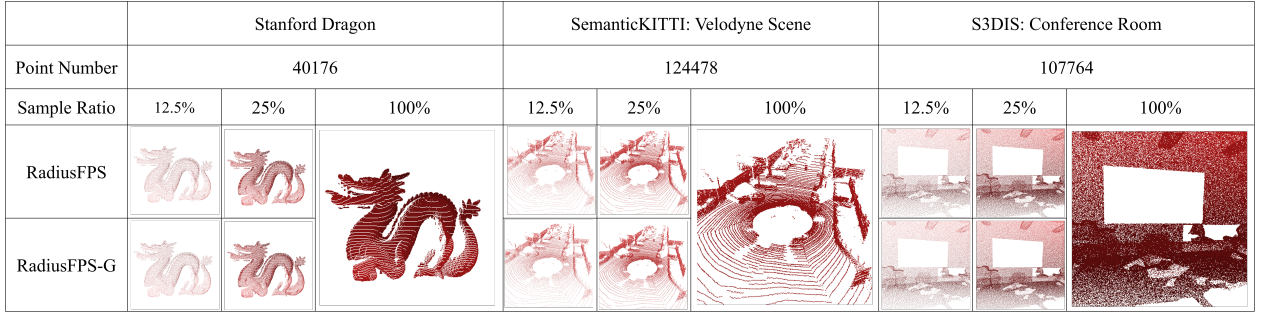}
    \caption{Visualization display of point cloud downsampling using our proposed RadiusFPS and RadiusFPS-G. We visualize the sampling results at 12.5\% and 25\% ratios against the original dense point clouds (100\%) across three diverse datasets. The visually consistent outputs demonstrate that our GPU-accelerated RadiusFPS-G preserves the spatial fidelity and structural integrity of the CPU-based RadiusFPS in these examples, even at highly sparse sampling rates.}
    \label{fig:vis}
\end{figure*}

\begin{figure}[t]
    \centering
    \includegraphics[width=0.8\columnwidth]{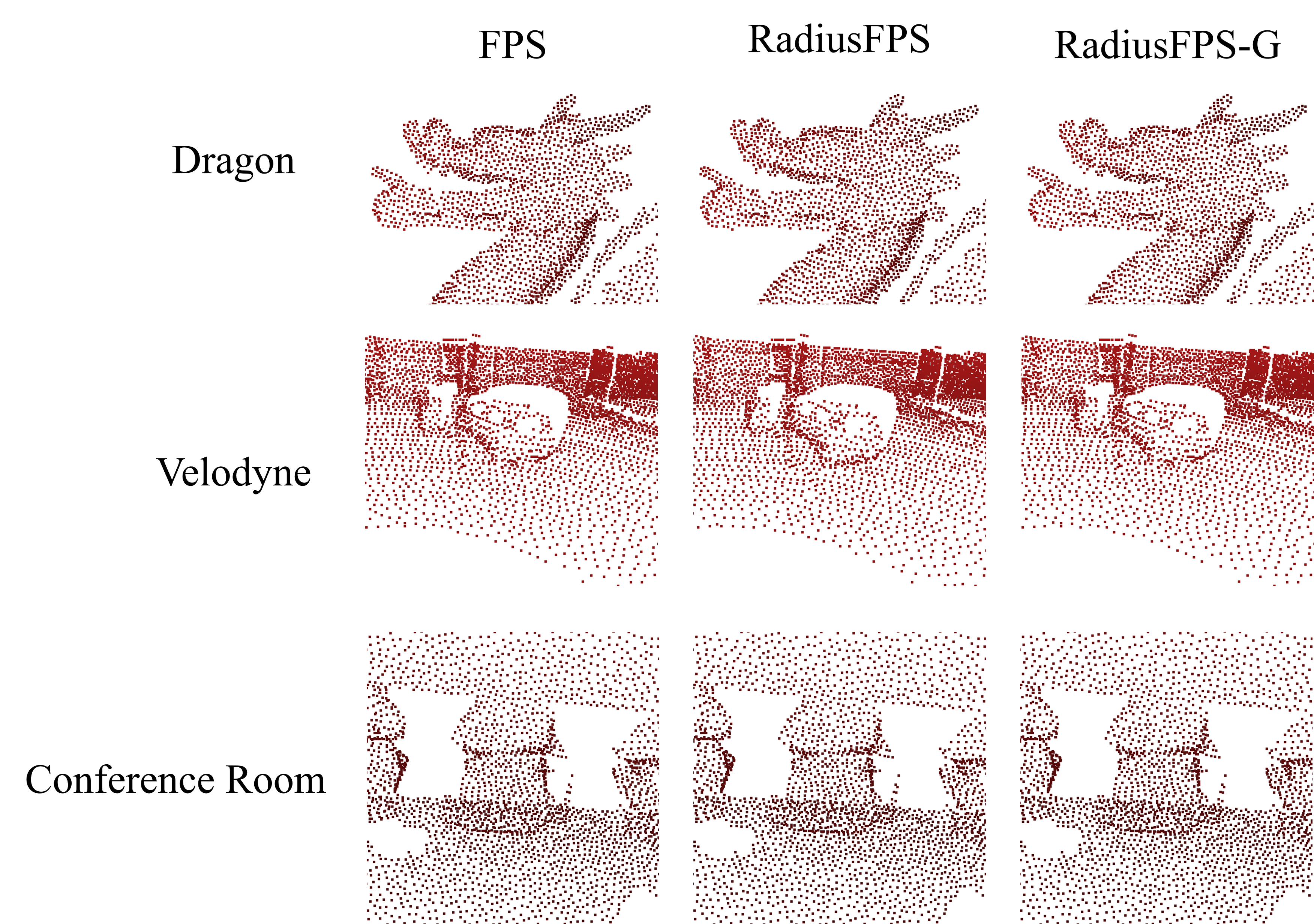}
    \caption{Close-up qualitative comparison of local geometric details. Zoomed-in views of the Stanford Dragon, SemanticKITTI Velodyne Scene, and S3DIS Conference Room datasets. The comparison illustrates that our proposed RadiusFPS and RadiusFPS-G (middle and right columns) effectively preserve fine-grained structures and maintain a highly uniform spatial distribution comparable to the FPS (left column).}
    \label{fig:vis_detail}
\end{figure}

\subsection{Quality Evaluation}
\subsubsection{End-to-End Segmentation Quality}
To isolate the impact of different sampling strategies on the final segmentation quality, we conduct a comprehensive quantitative evaluation, as summarized in Tab.\ref{tab:miou}. The performance is consistently measured by the mean Intersection over Union (mIoU) across three benchmark datasets (S3DIS, ScanNet, SemanticKITTI) utilizing both PointMetaBase and PointVector architectures.

As we can observe from the table, the results clearly demonstrate that basic heuristic downsampling methods, such as Random and Grid sampling, suffer from severe information loss. For instance, Random sampling on SemanticKITTI using PointMetaBase yields an mIoU of only 40.90\%, a significant drop from the 49.72\% achieved by the standard FPS (GPU) baseline. While EdgePC provides a more structured approach, it still falls short of the FPS standard across most setups. This highlights the inherent difficulty of accelerating point cloud sampling without compromising the spatial integrity required for dense prediction tasks.

In stark contrast, our proposed RadiusFPS and RadiusFPS-G consistently maintain highly competitive segmentation accuracy, improving the speed-quality trade-off. On the S3DIS and ScanNet datasets using the PointMetaBase architecture, RadiusFPS-G achieves mIoUs of 68.28\% and 71.21\%, respectively, slightly above the corresponding GPU-FPS baseline in these runs. On the large-scale SemanticKITTI dataset, the performance of RadiusFPS-G remains close to the GPU baseline (49.26\% vs. 49.72\% in PointMetaBase). Furthermore, when compared to alternative efficiency-oriented combinations like FPS + NPDU—which exhibits catastrophic accuracy degradation down to 26.64\% on SemanticKITTI with PointVector—our methods demonstrate strong robustness. This confirms that our optimizations preserve most critical geometric and contextual features during the downsampling process, ensuring that the downstream network receives high-quality representations despite the aggressive acceleration.

\subsubsection{Visualization Evaluation}
\label{sec:visualize}
To qualitatively evaluate the sampling quality of our proposed methods, we visualize the downsampled point clouds across different scales and densities. As illustrated in Fig.\ref{fig:vis}, we select three point cloud samples: Stanford Dragon\cite{stanfordscan} (over 40 thousand points), a Velodyne LiDAR point cloud scene (over 120 thousand points) from SemanticKITTI\cite{semantickitti}, and a complete conference room scan (over 1 million points) from S3DIS\cite{s3dis}. Each of these point clouds is sampled at two ratios: 12.5\% and 25\%. In the figure, both RadiusFPS and its GPU-accelerated variant, RadiusFPS-G, preserve the global geometric contours of the original dense point clouds, even under the sparse sampling ratio of 12.5\%. Notably, the visually consistent outputs between the two rows indicate that our GPU parallelization and fusion strategies preserve the spatial fidelity of the sampling algorithm in these examples. Furthermore, Fig.\ref{fig:vis_detail} provides zoomed-in views to assess local structural details. When compared to the standard FPS baseline, both RadiusFPS and RadiusFPS-G demonstrate highly consistent spatial distributions. They retain fine-grained features—such as the intricate shapes of the dragon and the sharp edges of the indoor furniture—without introducing irregular clustering or empty holes. Together, these visualizations indicate that our approaches achieve significant computational speedups while closely preserving the high-quality, uniform coverage characteristic of standard farthest point sampling.

\begin{figure*}[t]
    \centering
    \includegraphics[width=1\textwidth]{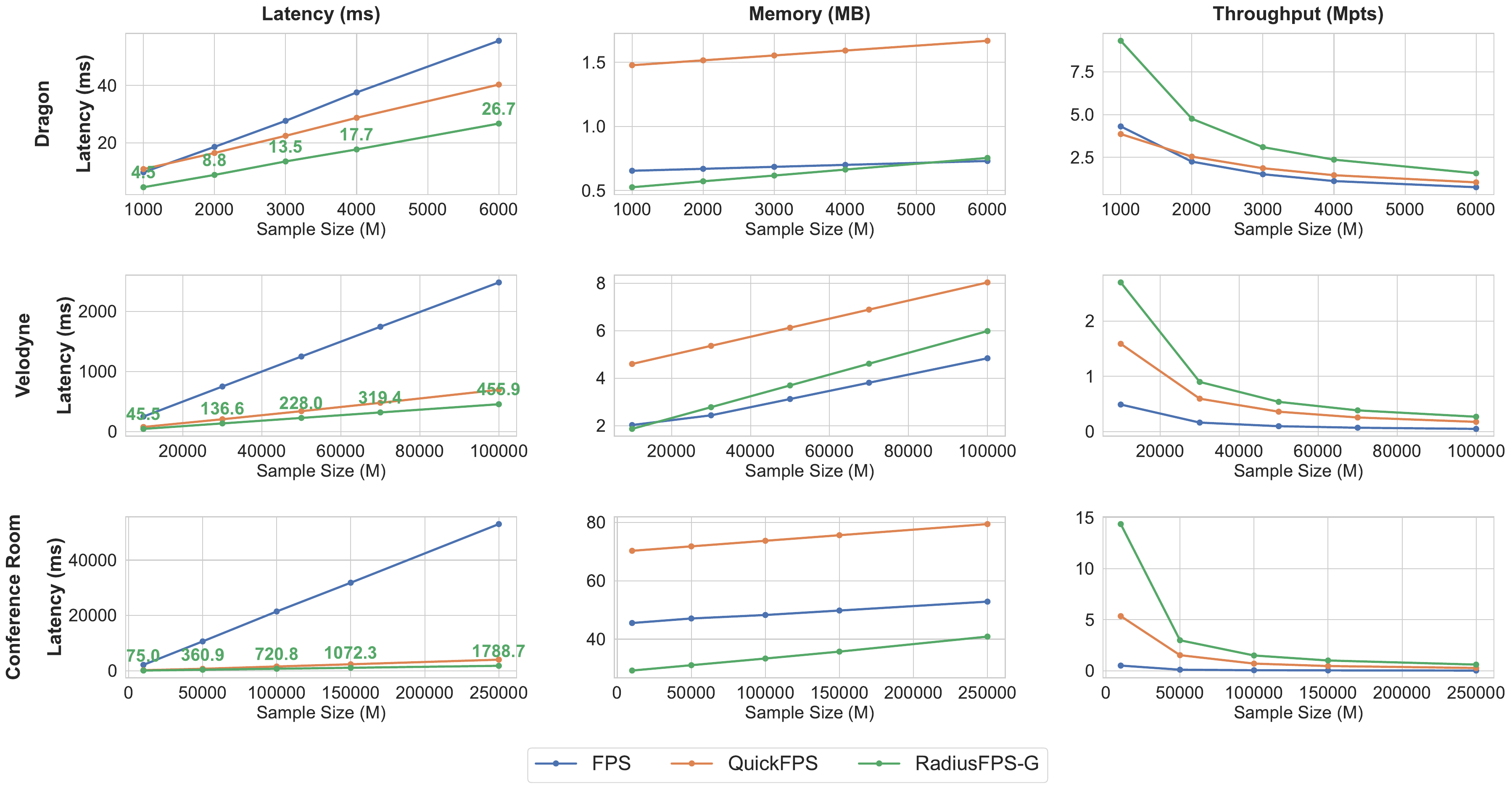}
    \caption{Performance scalability analysis across varying sample sizes. The line charts compare the Latency (ms), Memory consumption (MB), and Throughput (Mpts) of standard FPS, QuickFPS, and our proposed RadiusFPS-G on three representative datasets. Results demonstrate that RadiusFPS-G consistently achieves the lowest latency (annotated in green) and the highest throughput. Furthermore, it maintains a highly competitive memory footprint, scaling significantly better than QuickFPS on large-scale scenes.}
    \label{fig:gpuanalyze}
\end{figure*}

\begin{figure*}[h]
    \centering

    \subfloat[]{
    \includegraphics[width=\textwidth]{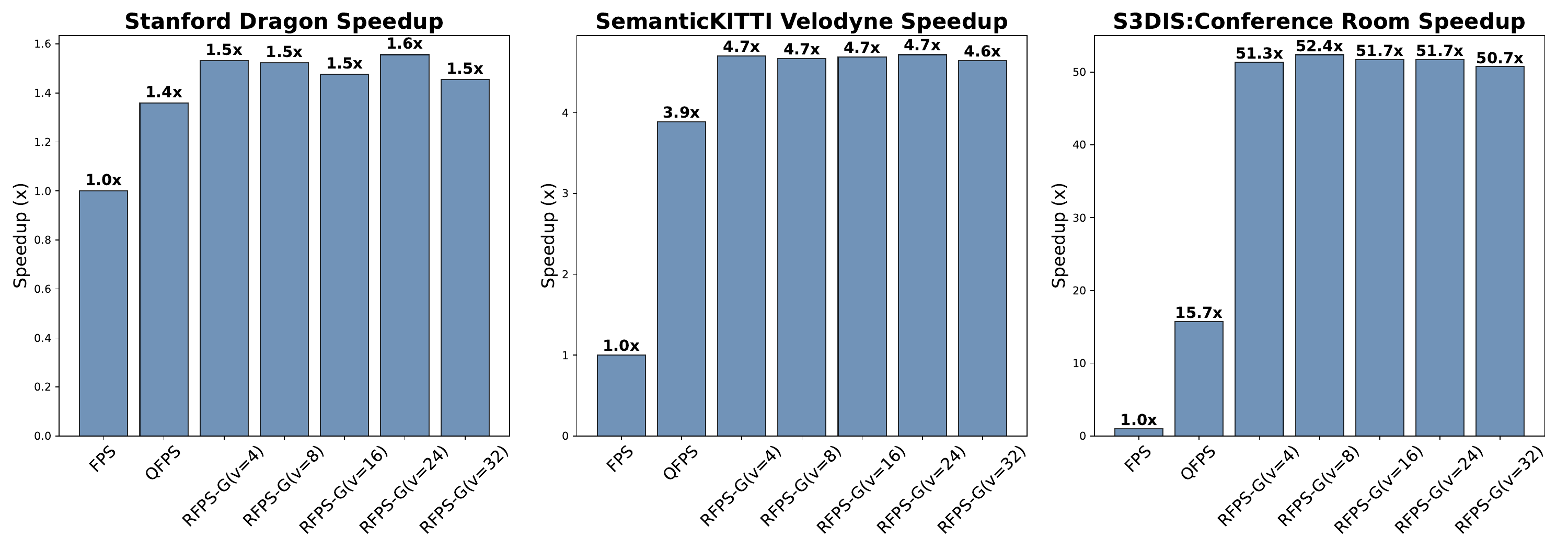}
    \label{fig:speedup_gpu}
    }
    \vspace{1em} 
    \subfloat[]{
    \includegraphics[width=\textwidth]{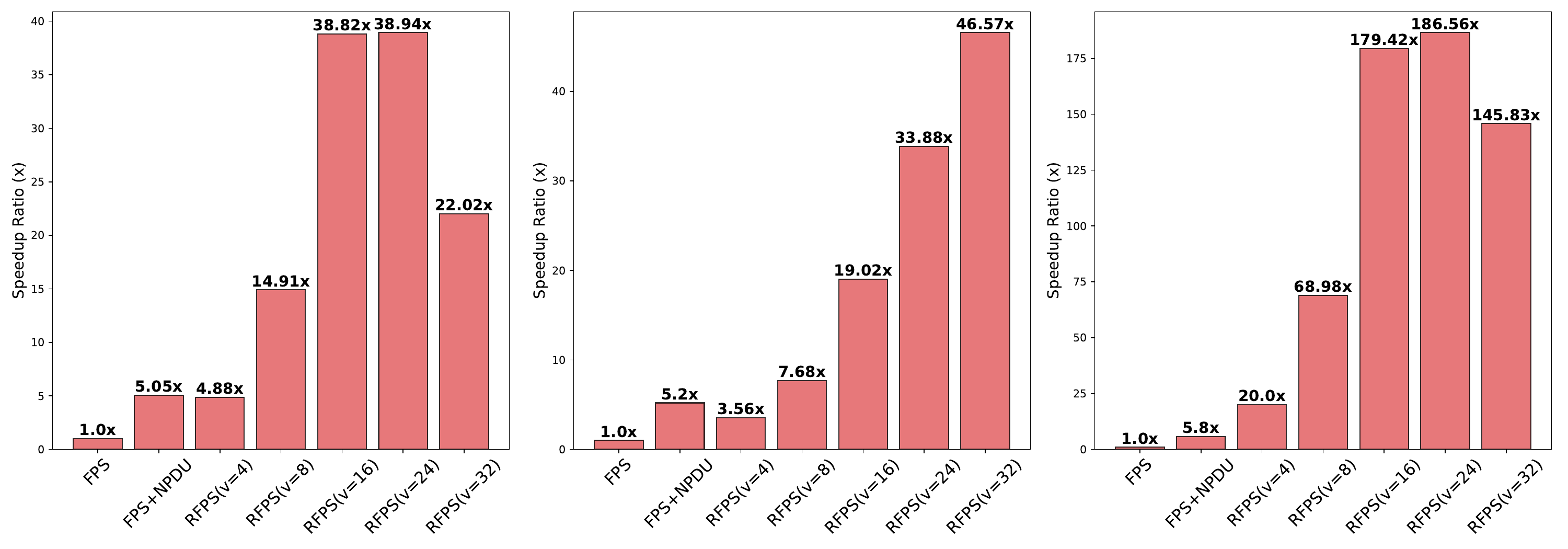}
    \label{fig:speedup_cpu}
    }
    \caption{Impact of varying voxel resolutions ($v$) on sampling speedup. The figures evaluate the performance scaling of (a) our GPU-accelerated RFPS-G against standard FPS and QuickFPS (QFPS), and (b) our CPU-based RFPS against FPS and FPS+NPDU, across three datasets. The results demonstrate that appropriately tuning the voxel resolution $v$ maximizes the performance gains, achieving peak speedups of up to 52.4$\times$ on GPU and 186.5$\times$ on CPU compared to the vanilla FPS baseline.}
\end{figure*}

\subsection{Scalability Studies}
To further evaluate RadiusFPS and RadiusFPS-G on large-scale point cloud processing, we conduct further tests of these algorithms on the three workloads mentioned in Sec.\ref{sec:visualize}.

\subsubsection{Computation Efficiency and Resource Consumption}
We first evaluate the computation efficiency and resource consumption by comparing RadiusFPS-G with FPS (GPU) and QuickFPS across the three samples. As illustrated in Fig.\ref{fig:gpuanalyze}, the performance is assessed using three metrics: latency, memory footprint, and throughput.

Regarding latency, FPS (GPU) exhibits a near-quadratic time complexity relative to the sample size, which becomes a severe bottleneck in dense or large-scale scenarios. For instance, on the SemanticKITTI dataset ($100,000$ points), the latency of vanilla FPS exceeds $2000$ ms, whereas RadiusFPS-G drastically reduces it to $455.9$ ms. This advantage is further magnified on the massive S3DIS Room dataset ($250,000$ points), where RadiusFPS-G maintains a remarkably stable execution time while vanilla FPS latency grows exponentially.

Crucially, this low latency is achieved without sacrificing memory efficiency. While alternative acceleration methods like QuickFPS incur substantial memory overhead (peaking at nearly $80$ MB for the Room dataset), RadiusFPS-G consumes approximately $40$ MB, maintaining a memory profile strictly comparable to vanilla FPS. Furthermore, RadiusFPS-G consistently delivers the highest throughput (Mpts) across all tested sample sizes, demonstrating its capability to sustain high-speed processing without introducing memory bottlenecks.

\subsubsection{Hardware Acceleration and Parameter Sensitivity}
We further investigate the speedup ratios across different computing architectures (GPU and CPU) and the algorithm's sensitivity to the voxel resolution parameter $v$. Fig.~\ref{fig:speedup_gpu} and Fig.~\ref{fig:speedup_cpu} detail the speedup comparisons against vanilla FPS and the CPU-optimized baseline, FPS+NPDU.

On the GPU backend (Fig.\ref{fig:speedup_gpu}), RadiusFPS-G demonstrates robust and consistent acceleration. The speedup scales effectively with scene complexity: while the Stanford Dragon yields a modest $1.5\times$ to $1.6\times$ improvement, the large-scale S3DIS Room dataset achieves an impressive $50\times$ (peaking at $52.4\times$) speedup over vanilla FPS. Notably, this GPU performance is largely insensitive to variations in the voxel resolution parameter $v$, indicating the robustness of our highly parallel voxelization and traversal kernels.

Conversely, the CPU implementation (Fig.\ref{fig:speedup_cpu}) reveals both unprecedented acceleration and distinct parameter sensitivity. RadiusFPS reaches a peak speedup of $186.56\times$ on the S3DIS conference room at $v=24$, orders of magnitude faster than the FPS+NPDU baseline (which achieves only a $5.8\times$ speedup). This massive gain highlights the effectiveness of our search-space pruning strategy in resource-constrained CPU environments.

However, the CPU speedup is highly sensitive to the chosen voxel resolution. For the Room dataset, the performance follows a non-monotonic trend, peaking at $v=24$ before declining to $145.83\times$ at $v=32$. This decline suggests that for certain spatial distributions, excessively fine voxels ($v=32$) introduce management overhead and cache misses from traversing numerous empty voxels, which eventually outweigh the benefits of pruning. In contrast, for the SemanticKITTI dataset, the speedup continues to increase up to $v=32$ (reaching $46.57\times$), likely benefiting from its specific, localized point distribution. These results emphasize that the optimal voxel resolution $v$ is a trade-off between indexing granularity and spatial distribution characteristics.

Fig.\ref{fig:vis} provides a qualitative comparison of the downsampled point clouds generated by our proposed RadiusFPS and RadiusFPS-G across three representative scenarios: a single object (Stanford Dragon), a large-scale outdoor LiDAR scene (SemanticKITTI Velodyne Scene), and a detailed indoor environment (S3DIS Conference Room), at sampling rates of 12.5\% and 25\%. Visually, the results for RadiusFPS-G are consistent with those of vanilla RadiusFPS, indicating that our parallel GPU implementation preserves the spatial properties of the sampling algorithm in these examples. Furthermore, across all datasets, both methods maintain the overall structural integrity and key geometric features (e.g., the intricate curves of the dragon and the layout of the furniture) even at the sparse 12.5\% sampling rate, demonstrating robust sampling capability across varying scales and environments.

\section{Conclusion}
In this paper, we introduce RadiusFPS, a farthest point sampling algorithm designed to improve computational efficiency while maintaining sampling quality. By employing a conservative radius pruning strategy, RadiusFPS preserves the FPS distance-update behavior under the same initialization and tie-breaking policy. Furthermore, we present RadiusFPS-G, a GPU-accelerated variant that integrates active-voxel packing and two fusion kernels to improve hardware utilization and memory access efficiency. Comprehensive evaluations on point cloud segmentation tasks demonstrate that both RadiusFPS and RadiusFPS-G achieve competitive accuracy while substantially reducing latency. Finally, by combining our heuristic approach with deep learning-based sampling techniques, we further improve inference speed with manageable accuracy changes, demonstrating potential for latency-sensitive, large-scale point cloud processing.

\bibliographystyle{unsrt}  
\bibliography{reference}  







\end{document}